\documentclass[10pt,twocolumn,letterpaper]{article}

\usepackage[pagenumbers]{cvpr}    

\usepackage{cuted}    
\usepackage{currfile} 
\usepackage{caption}  

\usepackage{multirow}

\usepackage{xspace}
\newcommand{\sys}{DeltaStyle\xspace}

\definecolor{cvprblue}{rgb}{0.21,0.49,0.74}
\usepackage[pagebackref,breaklinks,colorlinks,allcolors=cvprblue]{hyperref}

\def\confName{CVPR}
\def\confYear{2027}

\title{\sys: Residual Reuse for Efficient Streaming Video Stylization}

\author{Naoto Nishida\\
The University of Tokyo\\
{\tt\small nawta@g.ecc.u-tokyo.ac.jp}}

\begin{document}
\maketitle
\begin{strip}
\vspace{-0.9em}
\begin{center}
  \includegraphics[width=\textwidth]{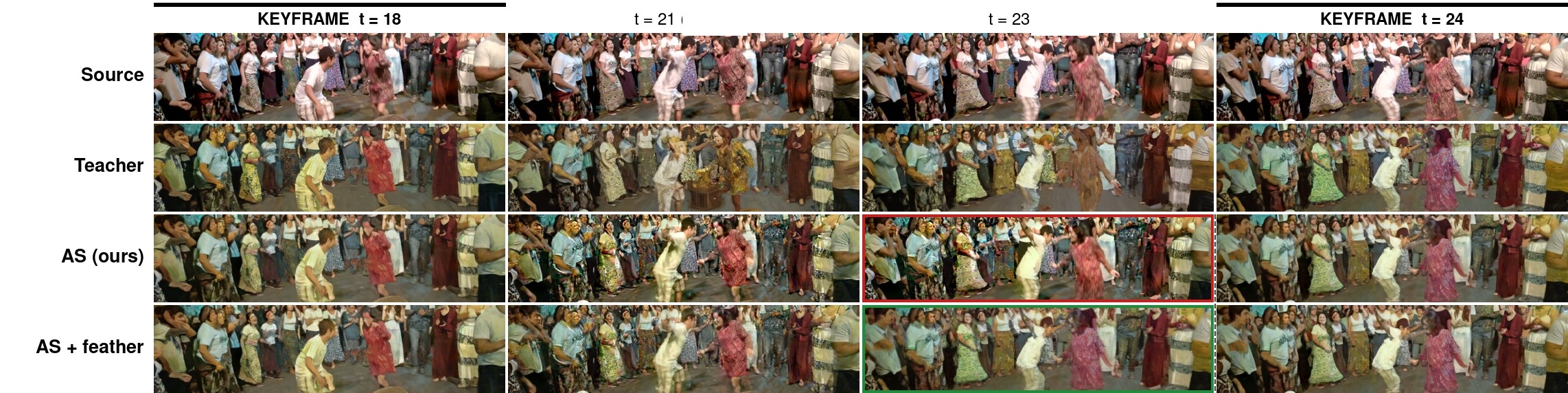}
\end{center}
\vspace{-1.5em}
\captionof{figure}{A frozen per-frame teacher evaluated only at
sparse keyframes ($t = 18, 24$). Anchor sampling (AS, third row) warps the keyframe residual to the frames between; the frame just before the next keyframe (red, $t = 23$) is furthest from its keyframe, and the output jumps at the boundary (dashed line); blending toward the next keyframe (AS + feather, bottom row, green; also ours) reduces the jump. One confirmatory-set clip (\emph{dancing}, van Gogh).}
\label{fig:teaser}
\end{strip}

\begin{abstract}
Per-frame diffusion stylization is the quality reference for video
stylization, but at about $222$\,ms per frame it is too slow for
streaming use. A cheaper alternative reuses a keyframe's
stylization residual---the teacher output minus the source
frame---warping it to the following frames with optical flow and
skipping the teacher between keyframes. We present a pre-registered
empirical measurement of residual reuse across frequency band,
transport vehicle, and operating point on held-out clip sets. A natural frequency-split hypothesis (chain the smooth low band,
anchor the detail-carrying high band) finds no statistically supported gain over anchoring the whole residual ($+0.003$ and $+0.028$ against a pre-registered $0.05$ threshold, $n = 10$ clips); the factorial attributes the gain over chained reuse mainly to anchoring the high band. An energy-matched control attributes about half of that gain
on ukiyo\_e ($45\%$) and most of it on van\_gogh ($89\%$) to
amplitude preservation alone, leaving a significant
amplitude-unexplained remainder only on ukiyo\_e ($+0.095$). Anchor sampling keeps about half of the teacher's style gain at an amortized $51$\,ms/frame (a quarter of the teacher's cost). Its temporal error has a specific anatomy: distance to the teacher grows with anchor age, and $24.5\%$ of the error mass sits on the one sixth of transitions that enter a keyframe ($9$ of $59$).
The transport vehicle is itself a trade: warping the keyframe image
instead of the residual retains more style but drifts in content and
snaps harder at boundaries, and chaining that image warp retains
still more style at a far larger temporal cost ($0.251$ vs.\
$0.142$). The anchor-over-chain ordering replicates under a second,
content-preserving teacher and on full-length sequences up to $91$ frames. At matched content preservation, FastBlend shows no detected style difference from anchor sampling and flickers less, so the advantage is causal streaming operation at roughly one fifth of the end-to-end cost. A lookahead feather blending each interval's last frames toward the next keyframe cuts the boundary jump by about half and raises retention to $0.62$ / $0.60$ on its design clips; on $10$ fresh clips the reduction confirms ($0.247 \to 0.108$), retention rises on ukiyo\_e ($+0.113$) and is flat on van\_gogh, at some content fidelity ($-0.076$ DINO-to-source) and no extra teacher calls; it is a buffered, near-live variant.
\end{abstract}
\section{Introduction}
\label{sec:intro}

Real-time diffusion pipelines such as
StreamDiffusion~\cite{kodaira2023streamdiffusion} make per-frame
stylization fast enough for live video, but a strong per-frame teacher
still costs $\approx$222\,ms/frame at $768{\times}448$ on a laptop GPU---too
slow for interactive rates. Reuse is the standard saving: stylize one keyframe every $K$ frames and transport
the stylization \emph{residual} (teacher output minus source frame)
to the frames in between with optical flow, at roughly a quarter of
the per-frame cost at $K{=}6$ (\cref{fig:teaser}).

A single transport rule is usually applied to the whole residual, yet the residual's frequency bands behave differently under transport:
iterated frame-by-frame warping (the chain) accumulates resampling
error in fine detail, while direct sampling from the keyframe anchor
keeps detail registered but lets smooth content drift.
This motivates a frequency-specific hypothesis---chain the low band,
anchor the high band---which we call Hybrid. Whether the split itself
carries any gain over anchoring the whole residual is an empirical
question that development-set results cannot settle, because the two
methods nearly tie there.

This paper is a pre-registered measurement study of residual
reuse---what carries the style signal, where the temporal error
concentrates, and what the resulting operating point costs and buys.
On a first held-out set, residual reuse with anchor transport keeps
measurably more of the teacher's style than chained or unsplit reuse
at matched keyframe budgets. On a second, fresh set, a confirmatory $2{\times}2$ factorial (band $\times$ transport) with a pre-registered decision rule finds the frequency-specific claim unsupported: the split matches whole-residual anchor sampling (AS) within noise, and the factorial localizes the gain over chaining predominantly to one factor, anchoring the high band. We therefore report anchor sampling as the supported mechanism. The anchor-over-chain ordering replicates across $20$ clips, under
a second content-preserving teacher (and a prompt-only one), on
sequences up to $91$ frames, and on a CSD-independent style metric;
the $\mathrm{Hybrid} \approx \mathrm{AS}$ verdict replicates on
ukiyo\_e and under the prompt-only teacher, while on van\_gogh CSD
gives a small positive difference and CLIP a small negative one. At
matched content preservation, an offline interpolation baseline
shows no detected style difference from anchor sampling and
flickers less, so anchor
sampling's contribution is a causal streaming operating point at
roughly one fifth of the end-to-end cost.

We also characterize where reuse's temporal error sits: distance to
the teacher grows with anchor age, $24.5\%$ of temporal-error mass
concentrates on the $9$ of $59$ boundary transitions per clip, and
jump size correlates with teacher residual drift ($r \approx 0.55$).
A fifteen-rater study dissociates the two axes: anchor sampling ties the offline interpolator on all axes ($48$--$49$ style, $41$--$49$ temporal, $44$--$44$ content vs FastBlend), outperforms image warp and lookahead feathering, and beats the half-resolution teacher on temporal stability ($116$--$1$) and on style ($95$--$21$).

Contributions: (i)~a measurement of anchor residual reuse for
causal streaming stylization, mapping its compute--style curve in
$K$ and content-matched trade-offs against offline keyframe
interpolation (\cref{sec:method,sec:comparison}); (ii)~a
pre-registered factorial test of frequency-specific transport,
finding no supported gain over anchor sampling and localizing the effect predominantly to the high band's transport (\cref{sec:ablation}); (iii)~an error anatomy of keyframe
reuse---anchor-age decay, boundary concentration, and an association
between the boundary jump and teacher drift (\cref{sec:anatomy});
and (iv)~a fifteen-rater evaluation comparing style, temporal stability, and content across baselines (\cref{sec:human}).
\section{Related work}
\label{sec:related}

\paragraph{Streaming diffusion.}
StreamDiffusion~\cite{kodaira2023streamdiffusion},
StreamDiffusionV2~\cite{feng2026streamdiffusionv2streamingdynamicinteractive},
and StreamV2V~\cite{liang2025lookingbackwardstreamingvideotovideo}
($20$\,FPS on an A100) run a generative model on (nearly) every frame;
\sys\ runs the teacher once per $K$ frames and reuses its output
residual.

\paragraph{Keyframe propagation.}
Classical and diffusion keyframe methods propagate stylized \emph{images} or learned fields: EbSynth~\cite{jamriska2019stylizingvideobyexample} uses guided patch synthesis; Rerender A Video~\cite{yang2023rerendervideozeroshottextguided} combines diffusion keyframes with EbSynth-style warping; FastBlend~\cite{duan2023fastblendpowerfulmodelfreetoolkit} blends keyframes without a model; Unified keyframe propagation~\cite{esser2022unifiedkeyframepropagationmodels} trains a propagator; StructuReiser~\cite{spetlik2025structureiser} runs a per-style network on every frame; FlowStyler~\cite{gong2025flowstyler} transports learned deformation and colour fields without keyframes. \sys\ instead transports the
output-space residual (teacher minus source), added to the untouched
source frame, with a chained-estimate fallback where flow is invalid
(\cref{sec:ablation}).

\paragraph{Delta propagation.}
VideoTimeTravel~\cite{kim2026videotimetravel} propagates a face re-aging delta map to adjacent frames via optical flow, but its delta comes from joint null-text optimization rather than a frozen teacher, uses adjacent-frame rather than composed anchor transport, and lacks a flow-confidence fallback.

\paragraph{Reuse inside the model and frequency views.}
Feature-caching and propagation methods such as DeepCache~\cite{ma2023deepcacheacceleratingdiffusionmodels}, TokenFlow~\cite{geyer2023tokenflowconsistentdiffusionfeatures}, and SmoothDiffusion-VE~\cite{munir2026smoothdiffusion} reuse features \emph{inside} the network while still evaluating the generator on every frame; \sys\ skips the generator between keyframes. Flow-warped consistency losses go back to Ruder et al.~\cite{ruder2016artisticstyletransfervideos}, and frequency decompositions appear inside training-free models such as FreeViS~\cite{xu2025freevistrainingfreevideostylization}; our factorial tests whether frequency should also govern residual transport (\cref{sec:ablation}).
\section{Method}
\label{sec:method}

\subsection{Frozen teacher}
A frozen one-step teacher stylizes a frame in two UNet
calls: SDXL-Turbo~\cite{sauer2023adversarialdiffusiondistillation,podell2023sdxlimprovinglatentdiffusion}
with a canny ControlNet~\cite{zhang2023addingconditionalcontroltexttoimage}
(scale $0.8$) and IP-Adapter~\cite{ye2023ipadaptertextcompatibleimage}
style injection at strength $0.65$, at $768{\times}448$. Applied to
every frame, with nearest-neighbour noise transport between frames
(the per-frame teacher), it defines the quality reference and cost
ceiling ($222.11$\,ms/frame; \cref{sec:comparison}); reuse methods are evaluated against this teacher's output, not ground truth.

\subsection{Residual reuse}
Every $K$ frames the teacher stylizes a keyframe $a$. Writing the fresh
keyframe residual as $r_a = T_a - X_a$, a non-keyframe output at frame
$t$ is $\hat{y}_t = X_t + \mathcal{T}(r_a)$, where $\mathcal{T}$
transports the fresh residual along RAFT-small~\cite{teed2020raftrecurrentallpairsfield}
optical flow estimated between adjacent source frames. Writing $b_t$
for the backward flow from $t$ to $t{-}1$ and $\mathcal{W}(\cdot, b_t)$
for bilinear grid sampling with border padding, the adjacent-frame
confidence map is
$c_t(p) = \mathbf{1}[p + b_t(p) \in \Omega \land e_t(p) \le 3\,\text{px}] \cdot
\exp(-(e_t(p)/1.5\,\text{px})^2)$,
where $\Omega$ is the frame domain and $e_t$ is the forward--backward
consistency error of the adjacent flows. Two transports are used:
	extbf{Chain (C):} the residual is warped frame by frame,
composing adjacent backward flows. \textbf{Anchor (A):} the composed
anchor flow $G_t$ from frame $t$ back to keyframe $a$ is updated
recursively as $G_t = b_t + \mathcal{W}(G_{t-1}, b_t)$ (with
$G_a = 0$), and composed confidence $C_t$ updates multiplicatively
the same way ($C_t = c_t \cdot \mathcal{W}(C_{t-1}, b_t)$,
$C_a = 1$). Pixels with $C_t(p) \ge 0.5$ and $p + G_t(p) \in \Omega$
sample the fresh residual directly from $a$ (anchor sampling, AS);
elsewhere the output falls back to the chained low-band estimate
(\cref{sec:factorial}).
At frame $0$ the displayed output is the teacher frame $T_0$
exactly; at later keyframes $a > 0$, the displayed keyframe residual
is updated as a confidence-weighted blend with $\lambda = 0.5$,
$\lambda c_a (\text{warped previous residual}) + (1 - \lambda c_a)(T_a - X_a)$,
and the same blend updates the chained low-band fallback, whereas
anchor sampling always transports the fresh residual $T_a - X_a$.

\begin{figure*}[tb]
\centering
\includegraphics[width=\textwidth]{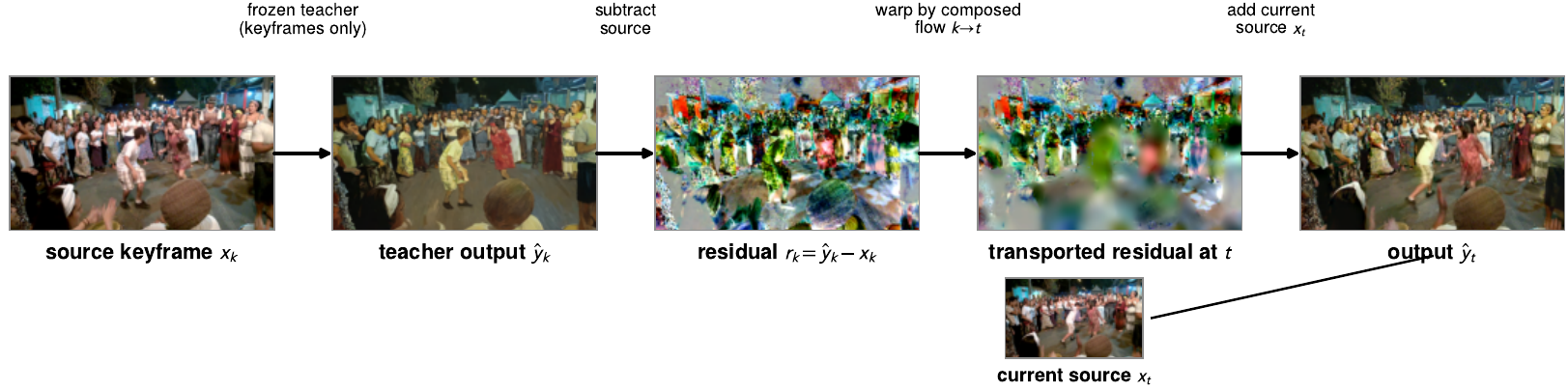}
\caption{Residual reuse. The teacher is evaluated only at
keyframes; its output residual is warped to each later frame by the
composed RAFT flow and added to that frame's source pixels, with a chained low-band fallback where the flow is invalid.}
\label{fig:pipeline}
\end{figure*}

\subsection{Frequency split and factorial design}
\label{sec:factorial}
The residual is split by a Gaussian filter with $\sigma = 8$ into a
low band $r^{L}$ and a high band $r^{H} = r - r^{L}$. The motivating
hypothesis is frequency-specific transport: iterated warping
accumulates error in fine detail but is stable for smooth content, so
the low band should be chained and the high band anchored,
\[
  \hat{y}_t = x_t + \mathcal{T}^{C}(r^{L}) + \mathcal{T}^{A}(r^{H})
  \quad \text{(Hybrid).}
\]
The $2{\times}2$ factorial over (low transport, high transport) gives
four cells---chain/chain, chain/anchor (the hybrid), anchor/chain,
and anchor/anchor (AS)---which
\cref{sec:ablation} uses to attribute the effect. Restricting the
chain/chain cell to the low band gives the chained low-band
estimate, the chained-reuse control. A causal exponential
moving average of the anchor-sampled residual (AS with EMA smoothing;
$\alpha \in \{0.3, 0.5, 0.7\}$ from development clips only)
serves as a smoothing null model.

\subsection{Cost model}
Per interval of $K$ frames the teacher runs once; the remaining cost
is flow estimation and warping, so measured cost falls from
$87.27$\,ms/frame at $K{=}3$ to $51.12$ at $K{=}6$ and $40.83$ at
$K{=}9$ (\cref{sec:comparison}): $K$ sets the compute--style
trade-off.
\section{Experiments}
\label{sec:experiments}

\subsection{Experimental setup}
\label{sec:setup}
DAVIS 2017 480p~\cite{ponttuset20182017davischallengevideo} (first $60$
frames per clip; full $62$--$91$ frames in \cref{sec:controls}).
Four disjoint clip sets are used: a development set for selection decisions, a first test set of $10$ held-out clips for main baselines, a fresh confirmatory set of $10$ clips (seed $2027$), and $10$ further fresh clips (seed $2028$) for expanded replication and held-out feather confirmation (clip lists in the supplementary).
Four styles are generated (watercolor, ukiyo\_e, van\_gogh,
ink\_sketch); the primary style metric uses ukiyo\_e and van\_gogh,
fixed on the development set before test outputs existed (all
comparisons pre-registered; supplementary).

\paragraph{Metrics.}
\textbf{CSD retention} (primary):
$(\mathrm{method} - \mathrm{source}) / (\mathrm{teacher} - \mathrm{source})$
on the CSD~\cite{somepalli2024measuringstylesimilaritydiffusion} ViT-L
style cosine to the style reference, on frames $5, 10, \dots, 55$, averaged within a clip, then over clips ($1$ means all teacher style gain is kept; absolute scores and denominators in the supplementary), read alongside raw CSD gain because ratios become unstable when a clip's teacher gap nears zero. \textbf{T\_LPIPS}:
flow-aligned
LPIPS~\cite{zhang2018unreasonableeffectivenessdeepfeatures} between
consecutive outputs (lower is less flicker); the \textbf{keyframe jump}
is T\_LPIPS on transitions into a keyframe. \textbf{Sharpness} is
Laplacian variance relative to the per-frame teacher, \textbf{LPIPS to
teacher} is distance to the per-frame teacher output, and content
preservation is DINOv2~\cite{oquab2024dinov2learningrobustvisual}
cosine to the source. Intervals are 95\% clip-cluster bootstraps
($2{,}000$ resamples, seed $0$, $n = 10$ clips), paired by clip;
confirmatory comparisons additionally report two-sided paired
permutation $p$-values ($10{,}000$ sign flips).

\paragraph{Methods compared.}
The per-frame teacher runs the full model on every frame (retention
$1$ by construction). Cheaper per-frame teachers spend less per frame
instead of skipping frames: the one-step teacher (one UNet call per
frame) and the half-resolution teacher (one UNet call at
$384{\times}224$, upsampled). Reuse methods comprise the chained
residual (previous frame's full-band residual warped forward), the
chained low-band baseline (also the fallback inside AS and Hybrid),
and sparse regeneration (low-confidence crops re-rendered once per
interval; supplementary). AS, Hybrid, and the factorial cells are
defined in \cref{sec:method}; image warp (IW) in \cref{sec:ablation}.

\subsection{Comparison with baselines}
\label{sec:comparison}
\begin{table*}[tb]
\centering
\scriptsize
\begin{tabular}{lccccc}
\toprule
 & & \multicolumn{2}{c}{$K{=}6$} & \multicolumn{2}{c}{$K{=}9$} \\
\cmidrule(lr){3-4}\cmidrule(lr){5-6}
Method & ms/frame ($K{=}6$ / $K{=}9$) & ukiyo\_e & van\_gogh & ukiyo\_e & van\_gogh \\
\midrule
One-step teacher & 137.76 & 0.90 [0.83, 0.97] & 0.69 [0.56, 0.79] & 0.90 [0.83, 0.97] & 0.69 [0.56, 0.79] \\
Half-res.\ teacher  & 91.02 & 0.50 [0.35, 0.64] & 0.44 [0.31, 0.59] & 0.50 [0.35, 0.64] & 0.44 [0.31, 0.59] \\
\midrule
Hybrid (ours) & 52.66 / 42.02 & \underline{0.59} [0.51, 0.69] & \textbf{0.69} [0.49, 0.86] & \underline{0.51} [0.41, 0.63] & \textbf{0.59} [0.39, 0.77] \\
Anchor sampling (proposed) & 51.12 / 40.83 & \underline{0.59} [0.51, 0.68] & \underline{0.64} [0.42, 0.82] & \underline{0.51} [0.42, 0.62] & \underline{0.54} [0.29, 0.74] \\
Sparse regen. (ours) & 83.89 / -- & \textbf{0.64} [0.57, 0.71] & 0.53 [0.28, 0.72] & \textbf{0.57} [0.48, 0.67] & 0.45 [0.18, 0.66] \\
Chained residual (ours)         & 49.84 / -- & 0.45 [0.36, 0.57] & 0.40 [0.20, 0.57] & 0.38 [0.27, 0.51] & 0.33 [0.15, 0.49] \\
Chained low-band (ours)     & 50.29 / -- & 0.41 [0.32, 0.50] & 0.26 [0.08, 0.41] & 0.32 [0.23, 0.44] & 0.19 [0.00, 0.35] \\
\bottomrule
\end{tabular}
\caption{CSD retention (mean [95\% CI], $n = 10$ clips) and amortized throughput. Teacher rows do not depend on $K$. Dashes (--) mark entries not benchmarked at that $K$; paired differences: supplementary material. The reuse rows are variants of our framework; anchor sampling is the proposed method. Bold marks the best value in each column among the reuse rows, underline the second best; teacher rows are references and stay unmarked.}
\label{tab:test-main}
\end{table*}

On the first test set (\cref{tab:test-main}; qualitative examples: supplementary), anchor-based reuse (Hybrid, AS) keeps more style than chained reuse at matched budgets: at $K{=}6$, Hybrid gains
$+0.18\,[0.16, 0.21]$ (ukiyo\_e) and $+0.42\,[0.36, 0.50]$
(van\_gogh) over the chained low-band baseline (paired differences
vs.\ chained residual: supplementary), flickers less than both
cheaper per-frame teachers ($0.137$ vs.\ $0.319$ / $0.224$) while
over-sharpening ($1.27\times$ teacher; \cref{sec:controls}), and
already nearly ties AS ($0.59$/$0.59$ ukiyo\_e; $0.69$/$0.64$
van\_gogh).

\paragraph{Compute--style trade-off.}
Timing is measured on an RTX 5080 Laptop GPU at $768{\times}448$, fp16, with CUDA synchronisation ($3$ warm-up frames, $3$ repeats; I/O excluded); per-method costs are collected in \cref{tab:timing} (AS $K{=}6$: $51.12 \pm 0.23$\,ms/frame). Amortized means hide a keyframe spike ($222.11$\,ms per keyframe vs.\ $20.09$\,ms measured for a non-keyframe AS frame: flow, warp, add); full sequences and isolated components are benchmarked separately, so the components do not recombine exactly into the amortized mean. The $51.12$\,ms/frame figure is the main table's RTX 5080 Laptop measurement; $45.1$\,ms/frame is the end-to-end RTX 4090 figure used for same-host baseline comparisons (\cref{sec:controls}). At $51.12$\,ms/frame the pipeline sustains $19.6$\,fps at $K{=}6$ ($24.5$ at $K{=}9$)---buffered or near-live streaming, with latency dominated by buffering.

For matched-content operating points on the confirmatory set ($K{=}6$), each output is blended toward the source ($\mathrm{source} + \alpha(\mathrm{method} - \mathrm{source})$) and all metrics recomputed (supplementary). Scaling FastBlend down to AS's
content level (DINO-to-source $0.829$, $\alpha = 0.756$) yields no
detected style-retention difference ($\text{FastBlend} - \mathrm{AS}$:
ukiyo\_e $+0.061$ $[-0.011, +0.134]$, van\_gogh $-0.021$
$[-0.317, +0.483]$) and less flicker (total T\_LPIPS $0.079$ vs.\
$0.142$; boundary $0.100$ vs.\ $0.228$). Scaling AS up cannot reach
FastBlend's retention at any $\alpha$ (maxima $0.506$ / $0.636$ vs.\
$0.838$ / $0.719$), and at FastBlend's content level AS flickers
worse (boundary $+0.249$ $[+0.226, +0.271]$). At matched content, AS
does not dominate the offline interpolator on any quality axis; its
advantage is causal streaming at $45.1$ vs.\ $\approx 243$\,ms/frame
end-to-end on the same GPU, whereas FastBlend requires both keyframes
of an interval.

\subsection{Ablation study}
\label{sec:ablation}
\begin{table*}[tb]
\centering
\footnotesize
\resizebox{\textwidth}{!}{%
\begin{tabular}{lccccc}
\toprule
Method ($K{=}6$) & ms/frame & ukiyo\_e & van\_gogh & T\_LPIPS & DINO-to-source \\
\midrule
Teacher (every frame) & 222.11 & 1.00 & 1.00 & 0.254 & -- \\
One-step teacher & 137.76 & 0.91 & 0.75 & 0.221 & -- \\
Half-res.\ teacher   & 91.02 & 0.54 & 0.42 & 0.293 & -- \\
\midrule
Chain, chain & 50.29 & 0.33 & 0.17 & \textbf{0.112} & \textbf{0.913} \\
Anchor, chain & -- & \underline{0.34} & 0.17 & \underline{0.114} & \underline{0.903} \\
Chain, anchor (Hybrid) & 52.66 & \textbf{0.51} & \textbf{0.47} & 0.138 & 0.845 \\
Anchor, anchor (AS) & 51.12 & \textbf{0.51} & \underline{0.44} & 0.142 & 0.829 \\
AS + EMA (exploratory)      & -- & 0.48 & 0.39 & 0.125 & 0.843 \\
\midrule
\multicolumn{6}{l}{\emph{Paired differences (mean [95\% CI]):}} \\
$\mathrm{Hybrid} - \mathrm{AS}$ & -- & $+0.003$ [$-0.013$, $+0.021$] & $+0.028$ [$-0.004$, $+0.063$] & $-0.003$ / $-0.003$ & $+0.015$ / $+0.016$ \\
$\mathrm{Hybrid} - (\mathrm{AS}+\mathrm{EMA})$ (exploratory) & -- & $+0.031$ [$+0.016$, $+0.047$] & $+0.085$ [$+0.050$, $+0.124$] & $+0.014$ / $+0.013$ & -- \\
\bottomrule
\end{tabular}}
\caption{Factorial results on the confirmatory set (fresh clips, $K{=}6$): CSD retention per style, mean T\_LPIPS, DINO-to-source, and amortized ms/frame where benchmarked. Cell labels give the (low-band, high-band) transports; bottom rows give paired differences (ukiyo\_e / van\_gogh where two values are shown). Bold marks the best value in each column among the reuse rows, underline the second best; teacher rows are references and stay unmarked.}
\label{tab:test2-factorial}
\end{table*}

\paragraph{Confirmatory factorial on the confirmatory set.}
The pre-registered decision rule: the claim passes
only if, on both styles, (1) Hybrid $-$ AS retention has a lower bound
above $0$ and a mean of at least $0.05$, (2) Hybrid $-$ AS DINO-to-source
has a lower bound of at least $-0.02$, and (3) Hybrid $-$ AS T\_LPIPS has
an upper bound of at most $+0.005$; the same rule applies to
Hybrid $-$ (AS with EMA smoothing), and if either comparison fails,
anchor sampling is reported as the supported mechanism.
The identity check (the chain/anchor cell must equal Hybrid, the
anchor/anchor cell AS) \textbf{passed bit-exactly} (mean $0.0$ on
all $40$ cells). Hybrid $-$ AS retention at $K{=}6$ is ukiyo\_e
$+0.003$ ($p = 0.77$) and van\_gogh $+0.028$
($p = 0.17$)---criterion (1) fails on both
styles; DINO-to-source ($+0.015$ / $+0.016$) and T\_LPIPS ($-0.003$
/ $-0.003$) pass their bounds. Hybrid $-$ (AS with EMA smoothing)
($\alpha = 0.7$ from the development set; exploratory):
retention $+0.031$ ($p = 0.002$) / $+0.085$
($p = 0.003$) favours Hybrid, but T\_LPIPS
($+0.014$ / $+0.013$) exceeds the $+0.005$ bound, so this comparison also does not confirm the claim. The frequency-specific claim is thus not confirmed, and anchor sampling is the supported mechanism.
Under the pre-registered rule, Hybrid's incremental benefit is not supported; this does not establish equivalence, and
the intervals above admit small positive differences.
The anchor-over-chain ordering replicates on the $10$ new clips
(ukiyo\_e per-frame retention $+0.204$ $[+0.132, +0.277]$); pooled
over $20$ clips, $\mathrm{AS} - \text{chained low-band baseline}$ is
$+0.189$ $[+0.150, +0.232]$ ($p < 0.001$, ukiyo\_e) and
$\mathrm{Hybrid} - \mathrm{AS}$ is $+0.001$ $[-0.017, +0.019]$. On
van\_gogh the per-frame estimator is unstable on the new clips (one
teacher gap is $-0.006$); a post-hoc ratio-of-means estimator
(exploratory) gives $\mathrm{Hybrid} - \mathrm{AS}$ $+0.038$
$[+0.019, +0.060]$ pooled (supplementary).
In the full factorial (\cref{tab:test2-factorial}), the paired main
effects are $+0.172$ ($p = 0.002$, ukiyo\_e) and
$+0.285$ ($p = 0.065$, van\_gogh; in raw CSD gain
the van\_gogh high-band effect is $+0.084$ $[+0.069, +0.100]$) for
anchoring the high band, versus $+0.002$ $[-0.013, +0.016]$
($p = 0.78$) and $-0.015$ ($p = 0.44$) for
anchoring the low band, with interaction at most $0.03$ in magnitude.
The factorial effect is thus predominantly the high band's
transport---strong on ukiyo\_e ($p = 0.002$), noisy in per-frame
retention on van\_gogh ($p = 0.065$) but clear in raw gain, while the low-band effect is undetected, which does not prove it is zero (the supplementary $\sigma$ sweep shows this is not a cutoff artifact): cells with an anchored high band retain $0.44$--$0.51$
versus $0.17$--$0.34$ when chained, so anchoring everything (AS)
already captures the band whose transport matters most. Binning confirmatory intervals into terciles by motion
magnitude and by fallback fraction shows that the anchor-over-chain
gain is positive in every bin for both styles and is about half as
large in the highest-fallback tercile---consistent with the
advantage living in the flow-valid region---while the feather
boundary reduction appears in every bin ($-0.08$ to $-0.12$) rather
than concentrating in high-motion intervals. Under anchor sampling ($K{=}6$), a mean of $26.8\%$ of non-keyframe pixels fall
back to the chained low-band baseline, rising with anchor age
(supplementary); all three anchored cells share
this fraction. Removing the fallback (invalid pixels take the source
frame) drops retention to $0.478$ / $0.336$ and worsens T\_LPIPS
(supplementary), yet stays far above the chained low-band baseline
($0.332$ / $0.174$): the fallback is load-bearing but does not
manufacture AS's advantage. All AS--chain pixel differences lie in the
flow-valid region by construction, which carries nearly all of the
whole-frame style advantage (valid-region raw gain $+0.053$
$[+0.041, +0.069]$ / $+0.078$ $[+0.059, +0.097]$ vs.\ $+0.054$ /
$+0.088$ whole-frame; supplementary).

\paragraph{Metric robustness (exploratory).}
In raw CSD gain on the confirmatory set ($K{=}6$),
$\mathrm{AS} - \text{chained low-band baseline}$ is $+0.049$
$[+0.039, +0.060]$ (ukiyo\_e) and $+0.080$ $[+0.064, +0.097]$
(van\_gogh), with high-band main effects $+0.048$ $[+0.039, +0.059]$
and $+0.084$ $[+0.069, +0.100]$ versus low-band effects $+0.001$ and $-0.004$ ($\approx 0$): the attribution is unchanged. For
$\mathrm{Hybrid} - \mathrm{AS}$, raw gain gives $+0.001$
$[-0.003, +0.005]$ (ukiyo\_e) and $+0.007$ $[+0.002, +0.011]$
(van\_gogh), ratio-of-means $+0.003$ $[-0.012, +0.020]$ and $+0.029$ $[+0.007, +0.059]$, and CLIP is negative in all four styles
($-0.007$ to $-0.016$ raw cosine, $10/10$ clips each): the raw-gain van\_gogh difference is positive but tiny, and inconsistent across styles, estimators, and metrics.

\paragraph{Transport vehicle.}
Image warp (IW) transports the stylized keyframe \emph{image} instead
of a residual at $K{=}6$. Stored IW differs from AS in flow, mask,
and keyframes as well as vehicle (direct $K{=}6$ flow/mask, raw
teacher keyframes): it retains more style ($0.66$ / $0.62$ vs.\
$0.51$ / $0.44$) and is calmer inside intervals ($0.090$ vs.\
$0.126$), but drifts in content (DINO-to-source $0.747$ vs.\ $0.829$)
and snaps harder at boundaries ($0.329$ vs.\ $0.228$). Matching all
three factors to AS (matched image warp; supplementary) keeps more
ukiyo\_e style ($0.624$; paired $+0.118$ [$+0.064, +0.194$];
van\_gogh $+0.032$, n.s.), sits closer to the teacher ($-0.023$
LPIPS), and flickers less ($-0.031$ total T\_LPIPS), while still
drifting in content (DINO $-0.070$) with a slightly larger boundary
jump ($+0.025$; raw teacher keyframes explain most of stored IW's
boundary excess). Chained image warp (propagating the previous output) retains more style ($0.73$ / $0.63$) at steep temporal cost (T\_LPIPS $0.251$ vs.\ $0.142$); remaining results use the residual vehicle (supplementary).

\subsection{Error anatomy}
\label{sec:anatomy}
\paragraph{Reading T\_LPIPS.}
Raw T\_LPIPS mixes temporal stability with style strength: the
chained low-band baseline is calmest ($0.112$) partly because it
carries the least style (retention $0.33$ / $0.17$), whereas the
per-frame teacher flickers most ($0.254$) at full style, so raw rankings alone do not order temporal quality. Splitting frame-to-frame change into the part
source motion explains (warp gain) and the rest: feather
lowers both warped and unwarped boundary differences ($-0.117$ and
$-0.115$) with unchanged warp gain ($+0.002$ )---
removing flicker rather than avoiding motion---whereas IW's boundary
excess carries a warp-gain loss of $-0.051$ 
(content jumps unexplained by source motion). Motion-normalised, AS's flicker is $1.84\times$ the source's own frame change overall and $2.90\times$ at boundaries; feather brings it to $1.40\times$ while \emph{raising} retention ($0.51$ / $0.44 \to 0.62$ / $0.60$) and beating the crossfade control on both.

\begin{figure*}[tb]
\centering
\includegraphics[width=\textwidth]{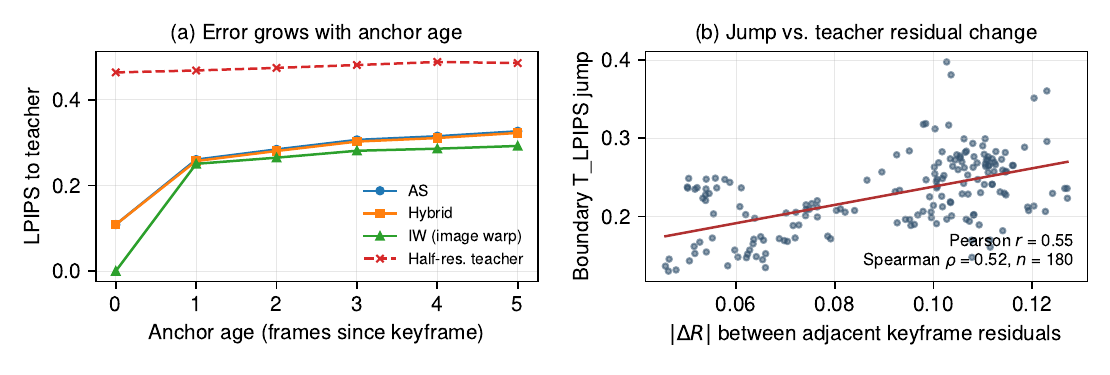}
\caption{Error anatomy on the confirmatory set ($K{=}6$). (a)~Distance to the per-frame teacher output grows with anchor age for every reuse method. (b)~Boundary jump against $\Delta R$, the teacher's residual change between keyframes.}
\label{fig:anatomy}
\end{figure*}

\paragraph{Anchor-age decay.}
Within an interval, the transported residual ages: for AS,
LPIPS-to-teacher grows monotonically with anchor age, from $0.109$ at
age $0$ to $0.327$ at age $5$ ($K{=}6$, confirmatory set;
\cref{fig:anatomy}(a)). The age-$0$ value comes from the confidence-weighted keyframe blend (\cref{sec:method}): frame~$0$ is exact ($0.000$), later keyframes average $0.121$, and a $\lambda$ sweep (supplementary) minimizes T\_LPIPS at $\lambda{=}0.5$.

\paragraph{Boundary concentration.}
At each boundary, the confidence-weighted keyframe blend pulls the
aged estimate toward the fresh teacher residual. On the confirmatory
set ($K{=}6$), boundary T\_LPIPS averages $0.228$ (AS) and $0.222$
(Hybrid)---$1.8\times$ the within-interval mean ($0.126$ /
$0.123$)---carrying $24.5\%$ of total T\_LPIPS mass on $9$ of $59$
transitions. A counterfactual decomposition of AS's boundary jump ($0.2275$)
gives source change $0.084$, transported-residual change $+0.039$
$[+0.030, +0.048]$, and teacher refresh $+0.105$ $[+0.092, +0.117]$
sequentially; refresh alone (source held fixed) already gives $0.210$, the largest single cause; the parts are non-additive (supplementary).
Per-frame baselines show no boundary spike (half-resolution teacher:
$0.293$ within, $0.296$ boundary).

\paragraph{Teacher drift.}
Across the $180$ confirmatory-set boundaries, jump size correlates
with teacher residual change $\Delta R$ between keyframes (Pearson $r = 0.55$, 95\% clip-cluster bootstrap CI $[0.13, 0.84]$; Spearman $\rho = 0.52$; \cref{fig:anatomy}(b)), associating $\approx 30\%$ of
jump variance; partial $r = 0.502$ (95\% CI $[0.08, 0.85]$)
controlling for interval-mean flow magnitude; the link is
correlational, not causal.

\subsection{Boundary feathering}
\label{sec:feather}
The anatomy suggests a direct remedy: for the last $F$ frames of
each interval, blend the anchor output toward the \emph{next}
keyframe's teacher frame, warped back through the adjacent flows
with a linear ramp; pixels where the backward warp is invalid keep
the anchor output. Feathering uses no extra teacher calls; a feathered frame waits for the next keyframe and its $222$\,ms teacher pass (the buffered, near-live setting).
$F = 3$ was selected on the design set as the ramp with the largest
retention gain among $F \in \{1, 2, 3\}$ while bringing the boundary
jump below within-interval levels. On the
confirmatory set at $K{=}6$ with $F{=}3$, the boundary jump falls from
$0.228$ to $0.110$, the
boundary share of T\_LPIPS drops from $24.5\%$ to $13.5\%$, total
T\_LPIPS falls from $0.142$ to $0.124$, and CSD retention
\emph{rises} from $0.51$ / $0.44$ to $0.62$ / $0.60$ (ukiyo\_e /
van\_gogh) (supplementary). Blending the next keyframe in early can also import its content errors: in the worst observed case, feather shows the teacher's subject replacement up to three frames early, where AS keeps the original subject (supplementary).

\paragraph{Confirmation on clips not used to design it.}
$F = 3$ was selected on the confirmatory (design) set and frozen
before running on the $10$ fresh seed-$2028$ clips: boundary
T\_LPIPS falls from $0.247$ to $0.108$ (paired $-0.139$
$[-0.158, -0.120]$, $p = 0.002$) and total T\_LPIPS by $-0.0175$
$[-0.0193, -0.0158]$, while ukiyo\_e retention rises by $+0.113$
 and van\_gogh style change is near zero (raw CSD
gain $+0.0006$ ). On the new clips the van\_gogh teacher gap is near zero on
several clips, leaving little style signal to add; the boundary
reduction confirms regardless. Feathering is thus a distinct operating point rather
than an unqualified improvement, trading content fidelity
(DINO-to-source $-0.076$  here; $0.829 \to 0.781$
on the confirmatory set) and lookahead latency for lower boundary
jumps (supplementary).

\paragraph{Lookahead control.}
A two-anchor crossfade control (same blend weights as feather, one
composed-flow warp) recovers most of the boundary reduction
(boundary T\_LPIPS $0.129$ vs.\ AS $0.228$; supplementary), but
paired $\text{feather} - \text{crossfade}$ intervals all exclude zero
(boundary $-0.019$ , total $-0.0055$
, retention $+0.049$ / $+0.113$, DINO $-0.023$
): lookahead alone does not explain the feather
gain.

\subsection{Controls and external baselines}
\label{sec:controls}
\paragraph{Sharpness-matched CSD control.}
Blurring Hybrid to match the teacher's clip$\times$style mean sharpness (Gaussian, $\sigma \in [0.30, 0.50]$) passes the pre-registered rule
(lower bound $> 0$ on both styles):
$\text{Hybrid-blur} - \text{chained low-band baseline} = +0.187$
$[+0.163, +0.216]$ (ukiyo\_e) and $+0.425$ $[+0.360, +0.492]$
(van\_gogh), with $\text{Hybrid-blur} - \text{Hybrid} \approx +0.003$
/ $+0.001$.

\paragraph{High-frequency energy control.}
Chaining also decays the high band's amplitude (RMS halves by anchor age $5$: $0.051 \to 0.025$), while anchor sampling holds it near-constant ($0.050$--$0.062$) and the low band stays flat.
Rescaling the chained estimate's high band per frame to match AS's
RMS (an oracle diagnostic, not a deployable method) separates two
effects within the $\mathrm{AS} - \text{chained low-band baseline}$
retention gap ($+0.174$ on ukiyo\_e and $+0.270$ on van\_gogh):
amplitude matching closes $45.4\%$  of the ukiyo\_e
gap, leaving an amplitude-unexplained remainder of $+0.095$
 ($54.6\%$), whereas on van\_gogh amplitude
matching closes $89.1\%$ and the remaining $+0.029$
 is not significant. These percentages are shares of the retention gap under this metric, not causal shares, and the remainder is not isolated structure (supplementary).

\paragraph{Second teacher.}
SD-Turbo (SD~2.1, one-step distilled, prompt-only
img2img without ControlNet or IP-Adapter), with strength calibrated
on the development set (documented deviation: the pre-set rule's
pick destroyed content at DINO $0.044$, so the next strength was
taken as primary: CSD gain $0.154$, DINO $0.319$). The
anchor-over-chain ordering replicates---raw CSD gain
$\mathrm{AS} - \text{chained low-band baseline}$ $+0.063$
 (ukiyo\_e) and $+0.072$ 
(van\_gogh)---as does $\mathrm{Hybrid} \approx \mathrm{AS}$ (raw
differences $-0.004$ / $-0.0004$, intervals spanning $0$), at all
three tested strengths (supplementary).

\paragraph{Second content-preserving teacher.}
The second content-preserving teacher raises the primary teacher's ControlNet scale from $0.8$ to $1.0$ and lowers img2img strength from $0.65$ to $0.50$, selected on development clips by a rule fixed in advance (strongest style with teacher
DINO-to-source $\ge 0.60$); its content preservation matches the
primary teacher's ($0.68$ vs.\ $0.64$; prompt-only teacher $0.33$).
On the
confirmatory set ($K{=}6$), the anchor-over-chain ordering
replicates: raw CSD gain $\mathrm{AS} - \text{chained low-band
baseline}$ is $+0.038$  (ukiyo\_e) and $+0.064$
 (van\_gogh), with similar temporal behavior (AS
T\_LPIPS $0.141$ total / $0.220$ boundary). $\mathrm{Hybrid} -
\mathrm{AS}$ is $+0.002$  on ukiyo\_e and $+0.006$
 on van\_gogh---a small positive difference of
about a tenth of the anchor-over-chain gap rather than an exact tie
(supplementary).

\paragraph{Replication on longer sequences.}
On full-length clips ($62$--$91$ frames; $176$ additional frames per style), the raw CSD gain of AS over the chained low-band baseline beyond frame $60$ is $+0.061$
 (ukiyo\_e) and $+0.097$ 
(van\_gogh), the boundary jump is unchanged ($0.227$ vs.\ $0.2275$
on the first $60$ frames), and the anchor-age decay keeps its shape
(AS above the baseline at every age; drift change consistent with
zero). Because DAVIS clips end by $91$ frames, this
tests moderate extensions rather than long videos.

\paragraph{CSD-independent style metric.}
On CLIP cosine to the style reference, AS beats the
chained low-band baseline on all four styles (paired $+0.031$ ukiyo\_e, $+0.057$ van\_gogh; other styles: supplementary), while IW and FastBlend score above AS; a VGG19 Gram metric is inconclusive (supplementary).

\begin{table}[t]
\centering
\scriptsize
\setlength{\tabcolsep}{4pt}
\begin{tabular}{llr}
\toprule
Method & Hardware & ms/frame \\
\midrule
Teacher (every frame) & RTX 5080 Laptop & 222.11 \\
Sparse regeneration (ours) & RTX 5080 Laptop & 83.89 \\
Anchor sampling (ours), $K{=}3/6/9$ & RTX 5080 Laptop & 87.27 / 51.12 / 40.83 \\
Hybrid (ours) & RTX 5080 Laptop & 52.66 \\
Chained low-band (ours) & RTX 5080 Laptop & 50.29 \\
\midrule
Anchor sampling, $K{=}6$ (ours) & RTX 4090 & 45.1 \\
FastBlend (end-to-end) & RTX 4090 & $\approx$243 \\
EbSynth, causal (sequential) & CPU & 1360 \\
EbSynth, causal (4 workers) & CPU & 620 \\
\bottomrule
\end{tabular}
\caption{Per-frame compute cost: amortized, under the paper's timing protocol (top) and same-host end-to-end (bottom). EbSynth: controlled run of \cref{sec:controls} (official CPU build).}
\label{tab:timing}
\vspace{-3ex}
\end{table}

\paragraph{External baselines.}
On the same confirmatory-set keyframes ($K{=}6$), FastBlend~\cite{duan2023fastblendpowerfulmodelfreetoolkit} (official interpolation mode) and EbSynth~\cite{jamriska2019stylizingvideobyexample} (official source build) show the same trade-off: more style than AS at lower content preservation; neither Pareto-dominates. Scaled toward the source to match AS's content preservation ($\alpha{=}0.68$), causal EbSynth's retention is unseparated from AS (pooled difference $+0.001$ $[-0.178, +0.272]$) and the boundary ordering reverses (lower T\_LPIPS and LPIPS-to-teacher for EbSynth). Per-frame costs are collected in \cref{tab:timing}: AS's advantage is GPU-rate causal operation.

\begin{table*}[tb]
\centering
\scriptsize
\begin{tabular}{lccc}
\toprule
Comparison (AS vs.) & Style & Temporal stability & Content \\
\midrule
Image warp & \textbf{77\% (43--13--64)} & \textbf{89\% (58--7--55)} & \textbf{86\% (51--8--61)} \\
AS with feathering & \textbf{67\% (53--26--41)} & \textbf{73\% (58--21--41)} & \textbf{74\% (55--19--46)} \\
FastBlend & 49\% (48--49--23) & 46\% (41--49--30) & 50\% (44--44--32) \\
Half-resolution teacher & \textbf{82\% (95--21--4)} & \textbf{99\% (116--1--3)} & \textbf{99\% (111--1--8)} \\
\bottomrule
\end{tabular}
\caption{Fifteen-rater pairwise study on Prolific ($32$ pairs per rater, ties allowed): share of non-tied judgments preferring AS (bold: majority for AS), with wins--losses--ties in parentheses; $120$ judgments per axis per comparison.}
\label{tab:human}
\end{table*}

\subsection{Human evaluation}
\label{sec:human}
Fifteen raters on Prolific give the judgments in \cref{tab:human}. AS was preferred to image warp and to AS with feathering on all three axes: feathering's boundary improvement does not appear as a perceptual preference. Against FastBlend the axes were even. Against the half-resolution teacher, AS was preferred overwhelmingly on temporal stability and content and also on style.

Implementation checks (supplementary): a bit-exact factorial, an independent rebuild of AS from the stored frames (mean $0.25/255$), and a cross-host audit.

\section{Conclusion}
\label{sec:conclusion}
\enlargethispage{\baselineskip}

Residual reuse makes a frozen per-frame teacher affordable for causal
streaming: at $K{=}6$ it keeps about half the teacher's style gain
at a quarter of the per-frame cost. A pre-registered factorial on fresh clips finds no supported gain for band-specific transport over anchor sampling; the gain localizes to the high band's transport, via amplitude preservation plus a remainder on ukiyo\_e. The ordering replicates across $20$ clips and a CSD-independent metric.
Practically, anchor sampling's advantage is causal operation at GPU
cost: at matched content, causal EbSynth ties on style and is temporally smoother at roughly two orders of magnitude more compute, while offline interpolation (FastBlend) shows no detected style difference and flickers less. Reuse's temporal error grows with anchor
age and concentrates at keyframe boundaries, where jump size correlates with the teacher's residual change between keyframes
($r = 0.55$; correlational). In a fifteen-rater study, viewers preferred AS to image warp and feathering on every axis, split evenly against FastBlend, and beat half-resolution generation on stability, content, and style.

\paragraph{Limitations.}
\looseness=-1
The primary metric is CSD on two styles (CLIP confirms it on all four). The prompt-only teacher lacks structure lock (DINO $0.33$ vs.\ $0.64$); the second content-preserving teacher ($0.68$) replicates it to $91$ frames; scene cuts and longer videos untested. Anchor reuse over-sharpens ($1.27\times$ teacher at $K{=}6$), ruled out by the sharpness-matched control.
\label{end-of-main-text}


\clearpage
\renewcommand{\thetable}{S\arabic{table}}
\renewcommand{\thefigure}{S\arabic{figure}}
\setcounter{table}{0}
\setcounter{figure}{0}
\begin{center}
    {\Large\bf Supplementary Material}\\[4pt]
    {\large \sys: Residual Reuse for Efficient Streaming Video Stylization}
\end{center}
\vspace{6pt}
\section{Overview and guide}
\label{sec:supp-overview}

The proposed configuration in the main paper is \textbf{anchor
sampling (AS)}: the whole stylization residual of a keyframe is
warped directly from that keyframe to each skipped frame, with a
chained low-band fallback where the flow is not trusted. Hybrid
(chained low band, anchored high band), image warp (IW),
feathering and the remaining rows of the main paper's tables are
controls and variants evaluated against AS, not separate
proposals. This supplementary material is the evidence record
for the main paper. Terms used throughout: \emph{retention} is
the per-frame mean of (method $-$ source) / (teacher $-$ source)
CSD gain; \emph{raw gain} is the unnormalised CSD gain;
\emph{chained low-band} (LOW-bK) is the chain/chain cell of the
transport factorial; the \emph{confirmatory set} is the ten
held-out DAVIS clips of the main paper's Sec.~4.3.
\Cref{tab:supp-definitions} defines every compared variant in
one line; the sections that follow use these names only.

\begin{table*}[t]
\centering
\small
\begin{tabular}{lp{0.70\textwidth}}
\toprule
Variant & Definition \\
\midrule
Teacher (B0NN) & The frozen stylizer run on every frame; the reference for gain and retention \\
Anchor sampling (AS) & The keyframe's whole residual warped from the keyframe by composed flow; chained low-band fallback on flow-invalid pixels; keyframes shown as a confidence-weighted blend ($\lambda = 0.5$) \\
Hybrid & Low band chained frame to frame, high band anchored at the keyframe; the factorial's chain/anchor cell \\
Factorial cells & The $2 \times 2$ design crosses each band's transport (anchored at the keyframe, or chained frame to frame): anchor/anchor $=$ AS; chain/anchor $=$ Hybrid; the other two cells chain the high band \\
Chained residual & The whole residual warped from the previous frame at every step, with no band split (Tab.~1 of the main paper); \cref{sec:supp-chains} reruns this transport as D0, with an exact teacher reset at keyframes \\
Chained low-band (LOW-bK) & The chain/chain cell of the factorial, run as a standalone baseline: the low band is chained frame to frame, flow-invalid pixels take a confidence-weighted Gaussian fill, and keyframes use AS's blended construction; it is also the fallback AS uses on flow-invalid pixels \\
Full-band chain (D0) & The whole residual chained with no fallback fill; resets to the exact teacher frame at keyframes \\
LOW-b & D0 whose flow-invalid pixels take the Gaussian fill instead of the warped value \\
Energy-matched chain (LOWHFM) & The chained low-band output rescaled per frame so its high-band amplitude matches AS's \\
Spectrum-matched chain (SPEC) & Oracle: the chained output given AS's own per-frame radial amplitude spectrum, phase unchanged; a diagnostic, not a runnable method \\
Image warp (IW) & The stylized keyframe \emph{image} (not the residual) warped to skipped frames; \emph{stored} uses direct $K$-frame flow and mask, \emph{matched} uses AS's composed flow, mask and blended keyframes \\
Crossfade & At a keyframe boundary, a linear blend of the two neighbouring anchors' propagated outputs over the feather window; like feathering it needs the next anchor before the boundary, but it blends propagated outputs rather than importing the teacher frame \\
Feathering (AS + feather) & AS whose frames within $3$ frames before a keyframe blend toward that keyframe's teacher output (lookahead; buffer $3$ frames) \\
AS + EMA & AS whose anchor residual is an exponential moving average of past keyframe residuals \\
Sparse regeneration & AS, except low-confidence crops are re-run through the teacher and composited in \\
AS, asynchronous & AS whose interval $j$ is displayed from keyframe $j{-}1$'s anchor while the teacher computes keyframe $j$ (\cref{sec:supp-ksweep}) \\
\bottomrule
\end{tabular}
\caption{The compared variants, defined once. Every method
above reuses the same keyframes, flow and confidence machinery
unless its line says otherwise.}
\label{tab:supp-definitions}
\end{table*}

\begin{table*}[t]
\centering
\small
\begin{tabular}{p{0.62\textwidth}p{0.31\textwidth}}
\toprule
Claim a reader may want to check & Where it is supported here \\
\midrule
Anchoring the high band drives the style gain & \Cref{sec:supp-r5} (factorial), \cref{sec:supp-bandfactorial,sec:supp-raftlarge}, \cref{fig:supp-forest} \\
Hybrid and AS are not separated on the primary retention estimator; on ukiyo\_e they are equivalent within $\pm 0.05$ retention, while on van\_gogh the raw-gain and ratio-of-means estimators separate them slightly in Hybrid's favour & \Cref{sec:supp-ci,sec:supp-r5,sec:supp-teacher2,sec:supp-teacher-cp} \\
The anchor beats the paper's chains and the full-band chains; a bicubic chain with Gaussian fill matches its raw gain only through exact teacher keyframes; under anchor sampling's keyframe construction it is the chained low-band baseline, which the anchor beats & \Cref{sec:supp-chains} \\
The result is not one teacher, set or metric & \Cref{sec:supp-new10,sec:supp-teacher2,sec:supp-clip,sec:supp-bandfactorial} \\
Errors concentrate at keyframe boundaries & \Cref{sec:supp-boundary-decomp,sec:supp-vehicle} \\
Feathering halves the boundary jump but costs content & \Cref{sec:supp-vehicle,sec:supp-feather-fail,sec:supp-human} \\
AS is a causal streaming operating point & \Cref{sec:supp-latency,sec:supp-basematrix,sec:supp-master} \\
The anchor's advantage holds at every refresh interval, and under the asynchronous policy & \Cref{sec:supp-ksweep} \\
The result is not one teacher family & \Cref{sec:supp-sanet} \\
External baselines trade quality against causality & \Cref{sec:supp-ebsynth,sec:supp-matched,sec:supp-streamdiffusion} \\
Human preferences partly disagree with the metrics & \Cref{sec:supp-human} \\
The implementation reproduces itself & \Cref{sec:supp-audit} \\
\bottomrule
\end{tabular}
\caption{Evidence map for the main paper's claims.}
\label{tab:supp-evmap}
\end{table*}

\section{Confidence intervals and paired differences}
\label{sec:supp-ci}

Table~1 of the main paper reports mean CSD retention on the first
test set. This section gives the corresponding 95\% confidence
intervals and the full paired-difference table. Intervals are
clip-cluster bootstrap intervals (2{,}000 resamples, $n = 10$ clips);
differences are paired by clip. The intervals are descriptive; no
significance-test family was registered. At $K{=}6$, Hybrid also
improves over the unsplit chained residual by $+0.14$ [$+0.10$,
$+0.17$] (ukiyo\_e) and $+0.29$ [$+0.24$, $+0.34$] (van\_gogh).

\begin{table*}[t]
\centering
\small
\begin{tabular}{lcccc}
\toprule
 & \multicolumn{2}{c}{$K{=}6$} & \multicolumn{2}{c}{$K{=}9$} \\
Method & ukiyo\_e & van\_gogh & ukiyo\_e & van\_gogh \\
\midrule
One-step teacher & 0.90 [0.83, 0.97] & 0.69 [0.56, 0.79] & 0.90 [0.83, 0.97] & 0.69 [0.56, 0.79] \\
Half-res.\ teacher & 0.50 [0.35, 0.64] & 0.44 [0.31, 0.59] & 0.50 [0.35, 0.64] & 0.44 [0.31, 0.59] \\
Hybrid & 0.59 [0.51, 0.69] & 0.69 [0.49, 0.86] & 0.51 [0.41, 0.63] & 0.59 [0.39, 0.77] \\
Anchor sampling & 0.59 [0.51, 0.68] & 0.64 [0.42, 0.82] & 0.51 [0.42, 0.62] & 0.54 [0.29, 0.74] \\
Sparse regen. & 0.64 [0.57, 0.71] & 0.53 [0.28, 0.72] & 0.57 [0.48, 0.67] & 0.45 [0.18, 0.66] \\
Chained residual & 0.45 [0.36, 0.57] & 0.40 [0.20, 0.57] & 0.38 [0.27, 0.51] & 0.33 [0.15, 0.49] \\
Chained low-band & 0.41 [0.32, 0.50] & 0.26 [0.08, 0.41] & 0.32 [0.23, 0.44] & 0.19 [0.00, 0.35] \\
\bottomrule
\end{tabular}
\caption{On the first test set, Hybrid and anchor sampling sit
together ($0.59$ / $0.59$ ukiyo\_e, $0.69$ / $0.64$ van\_gogh at
$K{=}6$), above both chained baselines and the half-resolution
teacher. CSD retention, mean [95\% CI], $n = 10$ clips. The
teacher rows do not depend on $K$.}
\label{tab:supp-test-ci}
\end{table*}

\begin{table*}[t]
\centering
\small
\begin{tabular}{lcccc}
\toprule
 & \multicolumn{2}{c}{$K{=}6$} & \multicolumn{2}{c}{$K{=}9$} \\
Difference & ukiyo\_e & van\_gogh & ukiyo\_e & van\_gogh \\
\midrule
Hybrid $-$ half-res.\ teacher & $+0.09$ [$-0.06$, $+0.24$] & $+0.25$ [$+0.08$, $+0.40$] & $+0.01$ [$-0.14$, $+0.16$] & $+0.15$ [$-0.01$, $+0.30$] \\
Hybrid $-$ one-step teacher & $-0.31$ [$-0.38$, $-0.24$] & $-0.00$ [$-0.10$, $+0.09$] & $-0.39$ [$-0.46$, $-0.31$] & $-0.11$ [$-0.21$, $+0.00$] \\
Hybrid $-$ chained low-band & $+0.18$ [$+0.16$, $+0.21$] & $+0.42$ [$+0.36$, $+0.50$] & $+0.19$ [$+0.16$, $+0.21$] & $+0.39$ [$+0.32$, $+0.47$] \\
Hybrid $-$ chained residual & $+0.14$ [$+0.10$, $+0.17$] & $+0.29$ [$+0.24$, $+0.34$] & $+0.13$ [$+0.10$, $+0.16$] & $+0.26$ [$+0.20$, $+0.31$] \\
Sparse regen.\ $-$ Hybrid & $+0.05$ [$+0.00$, $+0.09$] & $-0.15$ [$-0.26$, $-0.05$] & $+0.06$ [$+0.02$, $+0.11$] & $-0.13$ [$-0.26$, $-0.01$] \\
\bottomrule
\end{tabular}
\caption{Paired by clip, Hybrid separates from both chained
baselines on both styles at both $K$ (intervals exclude zero),
and does not separate from the one-step teacher on van\_gogh.
Paired differences in CSD retention on the first test set,
mean [95\% CI].}
\label{tab:supp-test-paired}
\end{table*}

\section{Pre-registration summary}
\label{sec:supp-prereg}

Clip sets: first test set (blackswan, train, bus, bmx-trees,
drift-turn, koala, sheep, dog-gooses, parkour, car-shadow);
confirmatory set, drawn by seed $2027$ (bike-packing, car-turn,
cat-girl, dancing, hike, kite-walk, mallard-fly, scooter-gray,
skate-park, stroller); third set, drawn by seed $2028$
(disc-jockey, drone, india, lady-running, paragliding, snowboard,
soapbox, tractor-sand, upside-down, varanus-cage).

Both test sets were pre-registered before the corresponding outputs
existed, and the two reported styles were fixed on the development
set before any test output existed; the decision rule for the
confirmatory factorial is stated in the main paper. Details held
here: the EMA grid $\alpha \in \{0.3, 0.5, 0.7\}$ was selected on
development clips only, with the documented deviation that no grid
value met the development-set rule (van\_gogh shortfall $0.002$),
so $\alpha = 0.7$ was used and the EMA comparison is exploratory.
The sharpness-matched control's pre-registered rule was
$\mathrm{Hybrid\text{-}blur} - \text{chained low-band}$ retention
lower bound $> 0$ on both styles.

\section{Remaining styles on the confirmatory set}
\label{sec:supp-styles}

The confirmatory set was run with two further styles, watercolor and
ink\_sketch, under the same protocol ($K{=}6$, same ten clips, same
evaluator). These styles place most of their energy in the low band,
so the factorial analysis of the main paper predicts that anchoring
the high band should matter less. The results match that prediction:
the chained low-band estimate is at least as close to the teacher as
anchor sampling on both metrics, and Hybrid tracks anchor
sampling, replicating the null reported for the two main styles.
CSD retention and DINO-to-source were not computed for these two
styles; the metrics below are LPIPS to the teacher's output
(mean over frames, excluding each clip's first frame) and
T\_LPIPS (mean over consecutive-frame transitions; boundary
transitions are the transitions into a keyframe frame).

\begin{table*}[t]
\centering
\small
\begin{tabular}{lccccc}
\toprule
 & & \multicolumn{3}{c}{T\_LPIPS} \\
\cmidrule(lr){3-5}
Style / Method & LPIPS & total & internal & boundary \\
\midrule
\multicolumn{5}{l}{watercolor} \\
Teacher (every frame) & 0.000 & 0.261 & 0.261 & 0.261 \\
Chained low-band & 0.272 & 0.112 & 0.100 & 0.180 \\
Anchor sampling & 0.283 & 0.143 & 0.128 & 0.229 \\
Hybrid & 0.280 & 0.139 & 0.125 & 0.222 \\
\midrule
\multicolumn{5}{l}{ink\_sketch} \\
Teacher (every frame) & 0.000 & 0.243 & 0.244 & 0.243 \\
Chained low-band & 0.257 & 0.113 & 0.102 & 0.174 \\
Anchor sampling & 0.265 & 0.144 & 0.130 & 0.221 \\
Hybrid & 0.263 & 0.141 & 0.127 & 0.216 \\
\bottomrule
\end{tabular}
\caption{Remaining styles on the confirmatory set ($K{=}6$,
10 clips). The teacher's LPIPS to itself is zero by construction;
its T\_LPIPS is the temporal variation of the per-frame teacher
itself.}
\label{tab:supp-extra}
\end{table*}

Paired by clip, anchor sampling minus the chained low-band estimate
is $+0.011$ [$+0.009$, $+0.013$] LPIPS and $+0.031$ [$+0.027$,
$+0.035$] T\_LPIPS for watercolor, and $+0.009$ [$+0.002$, $+0.014$]
LPIPS and $+0.032$ [$+0.028$, $+0.036$] T\_LPIPS for ink\_sketch
(95\% clip-level bootstrap intervals, 10{,}000 resamples). Positive
values favor the chained estimate; both intervals exclude zero for
both styles and both metrics.

\begin{table*}[t]
\centering
\small
\begin{tabular}{llcccc}
\toprule
 & & \multicolumn{2}{c}{LPIPS} & \multicolumn{2}{c}{T\_LPIPS} \\
\cmidrule(lr){3-4}\cmidrule(lr){5-6}
Style & Clip & AS & chained & AS & chained \\
\midrule
watercolor & bike-packing & 0.231 & 0.222 & 0.113 & 0.086 \\
 & car-turn & 0.276 & 0.261 & 0.130 & 0.096 \\
 & cat-girl & 0.277 & 0.264 & 0.144 & 0.112 \\
 & dancing & 0.294 & 0.278 & 0.193 & 0.169 \\
 & hike & 0.318 & 0.305 & 0.158 & 0.115 \\
 & kite-walk & 0.326 & 0.320 & 0.131 & 0.090 \\
 & mallard-fly & 0.376 & 0.367 & 0.188 & 0.155 \\
 & scooter-gray & 0.228 & 0.225 & 0.118 & 0.092 \\
 & skate-park & 0.193 & 0.180 & 0.089 & 0.065 \\
 & stroller & 0.314 & 0.299 & 0.165 & 0.141 \\
\midrule
ink\_sketch & bike-packing & 0.211 & 0.202 & 0.113 & 0.085 \\
 & car-turn & 0.253 & 0.238 & 0.131 & 0.098 \\
 & cat-girl & 0.265 & 0.251 & 0.145 & 0.113 \\
 & dancing & 0.286 & 0.271 & 0.196 & 0.171 \\
 & hike & 0.294 & 0.279 & 0.159 & 0.115 \\
 & kite-walk & 0.313 & 0.331 & 0.129 & 0.089 \\
 & mallard-fly & 0.348 & 0.340 & 0.195 & 0.160 \\
 & scooter-gray & 0.218 & 0.218 & 0.121 & 0.092 \\
 & skate-park & 0.163 & 0.153 & 0.086 & 0.063 \\
 & stroller & 0.303 & 0.283 & 0.167 & 0.141 \\
\bottomrule
\end{tabular}
\caption{Per-clip values for the remaining styles ($K{=}6$).
AS: anchor sampling; chained: chained low-band estimate. The chained
estimate has the lower T\_LPIPS on all 20 clip--style units and the
lower LPIPS on 18 of 20.}
\label{tab:supp-extra-clips}
\end{table*}

\section{Frequency-split scale sweep}
\label{sec:supp-sigma}

To check whether the factorial verdict depends on the Gaussian cutoff
$\sigma = 8$, we sweep $\sigma \in \{4, 8, 16, 32\}$ on the
confirmatory set ($K{=}6$, \cref{tab:supp-sigma}). Across these
scales, AS retention ranges $0.498$--$0.507$ (ukiyo\_e) and
$0.418$--$0.461$ (van\_gogh); $\sigma = 4$ has the best temporal
scores but the worst retention. Overall, $\sigma = 8$ is not
dominated, and no $\sigma$ makes Hybrid's advantage consistent across
styles.

\begin{table*}[t]
\centering
\small
\begin{tabular}{ccc}
\toprule
 & \multicolumn{2}{c}{$\mathrm{Hybrid} - \mathrm{AS}$ CSD retention} \\
\cmidrule(lr){2-3}
$\sigma$ & ukiyo\_e & van\_gogh \\
\midrule
$4$  & $+0.006$ & $-0.015$ \\
$8$  & $+0.003$ & $+0.028$ \\
$16$ & $-0.003$ & $+0.035$ \\
$32$ & $-0.005$ & $+0.009$ \\
\bottomrule
\end{tabular}
\caption{Frequency-split scale sweep on the confirmatory set
($K{=}6$, $n = 10$ clips): paired $\mathrm{Hybrid} - \mathrm{AS}$ CSD
retention for $\sigma \in \{4, 8, 16, 32\}$. AS retention ranges
$0.498$--$0.507$ (ukiyo\_e) and $0.418$--$0.461$ (van\_gogh);
$\sigma = 4$ has the best temporal scores but the worst retention.
$\sigma = 8$ is not dominated, and no $\sigma$ makes Hybrid's
advantage consistent across styles.}
\label{tab:supp-sigma}
\end{table*}

\section{Absolute CSD scores, fallback rates, and region decomposition}
\label{sec:supp-csd-abs}

On the confirmatory set ($K{=}6$, ukiyo\_e / van\_gogh), absolute CSD
style cosines are: source $-0.0497$ / $-0.0043$, per-frame teacher
$0.2270$ / $0.2209$, AS $0.0907$ / $0.1513$, and Hybrid $0.0915$ /
$0.1578$. The per-clip retention denominator
$(\mathrm{teacher} - \mathrm{source})$ is $0.277 \pm 0.041$
(min $0.203$) for ukiyo\_e and $0.225 \pm 0.121$ (min $0.054$) for
van\_gogh; van\_gogh's more variable denominator matches its wider
retention confidence intervals.

Under anchor sampling on the confirmatory set ($K{=}6$), the
fallback fraction rises with anchor age: $12.0\%$, $20.6\%$,
$28.0\%$, $34.3\%$, and $39.4\%$ at ages $1$--$5$ ($22.4\%$ over
all frames). Removing the chained low-band fallback from AS
(invalid pixels take the source frame) worsens T\_LPIPS to
$0.155$ total, $0.138$ internal, and $0.254$ boundary (vs.\ AS's
$0.142$ / $0.126$ / $0.228$), alongside the retention drop
reported in the main text.

\Cref{tab:supp-region} breaks down the AS vs.\ chained low-band
comparison into the flow-valid region ($73.2\%$ of skipped-frame
pixels) and the fallback region ($26.8\%$, where AS and the
chained low-band estimate are identical by construction). On
valid-region LPIPS-to-teacher AS sits slightly farther from the
teacher than the chained low-band estimate ($0.2996$ vs.\
$0.2878$ pooled across both styles, $+0.012$ [$+0.007, +0.017]$).

\begin{table*}[t]
\centering
\small
\begin{tabular}{lcccc}
\toprule
 & \multicolumn{2}{c}{Raw CSD gain ($\mathrm{AS} - \text{chained low-band}$)} & \multicolumn{2}{c}{Valid-region LPIPS-to-teacher} \\
\cmidrule(lr){2-3}\cmidrule(lr){4-5}
Style & flow-valid ($73.2\%$) & whole-frame & AS & chained low-band \\
\midrule
ukiyo\_e  & $+0.053$ [$+0.041$, $+0.069$] & $+0.054$ & $0.2992$ & $0.2887$ \\
van\_gogh & $+0.078$ [$+0.059$, $+0.097$] & $+0.088$ & $0.2999$ & $0.2868$ \\
\bottomrule
\end{tabular}
\caption{Region decomposition on the confirmatory set ($K{=}6$,
non-keyframe frames). The fallback region ($26.8\%$ of pixels) is
identical between AS and the chained low-band estimate by
construction.}
\label{tab:supp-region}
\end{table*}

\section{Pure-teacher keyframe control and blend-weight sweep}
\label{sec:supp-pure-kf}

To isolate the effect of the confidence-weighted keyframe residual
blend (main paper, Sec.~3), we evaluated a variant of AS whose
keyframes are the pure teacher frames ($\lambda{=}0$) on the
confirmatory set ($K{=}6$). This variant eliminates the age-$0$
LPIPS-to-teacher offset ($0.000$ vs.\ AS's $0.109$) and raises CSD
retention to $0.542$ (ukiyo\_e, \cref{tab:supp-lambda}) /
$0.504$ (van\_gogh, control run), but enlarges
the boundary jump to $0.340$ and worsens total T\_LPIPS to $0.150$,
compared to AS's $0.228$ boundary jump and $0.142$ total T\_LPIPS.
\Cref{tab:supp-lambda} reports the full sweep over the keyframe
blend weight $\lambda \in \{0, 0.25, 0.5, 0.75, 1\}$: ukiyo\_e
retention falls monotonically as $\lambda$ increases, the boundary
jump is smallest at $\lambda{=}0.75$ ($0.196$), and total T\_LPIPS is
minimized at $\lambda{=}0.5$ ($0.1415$).

\begin{table*}[t]
\centering
\small
\begin{tabular}{ccccc}
\toprule
$\lambda$ & ukiyo\_e retention & boundary T\_LPIPS & total T\_LPIPS & DINO-to-source \\
\midrule
$0.00$ & $0.542$ & $0.3400$ & $0.1499$ & $0.815$ \\
$0.25$ & $0.530$ & $0.2900$ & $0.1450$ & $0.821$ \\
$0.50$ & $0.506$ & $0.2275$ & $0.1415$ & $0.829$ \\
$0.75$ & $0.484$ & $0.1960$ & $0.1452$ & $0.834$ \\
$1.00$ & $0.467$ & $0.2100$ & $0.1573$ & $0.833$ \\
\bottomrule
\end{tabular}
\caption{Keyframe residual blend-weight sweep on the confirmatory set
($K{=}6$, $n = 10$ clips). $\lambda = 0.5$ minimizes total
T\_LPIPS across the grid.}
\label{tab:supp-lambda}
\end{table*}

\section{Counterfactual boundary decomposition}
\label{sec:supp-boundary-decomp}

\Cref{tab:supp-boundary} decomposes AS's boundary T\_LPIPS ($0.2275$
$[0.2006, 0.2540]$ on the confirmatory set at $K{=}6$) into its
contributing transitions: source-frame change accounts for
$0.0838$, transported-residual change adds $+0.039$ $[+0.030,
+0.048]$ (reaching $0.1227$ without a keyframe refresh), and
teacher refresh alone with the source held fixed produces
$0.2102$.

\begin{table*}[t]
\centering
\small
\begin{tabular}{lc}
\toprule
Counterfactual transition ($K{=}6$) & Boundary T\_LPIPS \\
\midrule
Nothing changes (sanity check) & $0.0000$ \\
Source change only & $0.0838$ \\
Source + transported-residual change (no refresh) & $0.1227$ \\
Teacher refresh only (source held fixed) & $0.2102$ \\
Actual AS boundary transition & $0.2275$ [$0.2006$, $0.2540$] \\
\bottomrule
\end{tabular}
\caption{Counterfactual decomposition of AS boundary T\_LPIPS on the
confirmatory set ($K{=}6$, $180$ boundaries).}
\label{tab:supp-boundary}
\end{table*}

\section{Expanded 20-clip replication and held-out feather confirmation}
\label{sec:supp-new10}

To test replication beyond the $10$ confirmatory clips, we drew $10$
further DAVIS clips by seed $2028$ from the remaining $41$-clip pool
(disc-jockey, drone, india, lady-running, paragliding, snowboard,
soapbox, tractor-sand, upside-down, varanus-cage) and ran the same
$K{=}6$ protocol (\cref{tab:supp-new10}). On van\_gogh, the
per-frame retention estimator is unstable on the new clips because
one clip (upside-down) has a negative teacher style gap ($-0.006$);
the table's last column therefore uses an exploratory post-hoc
ratio-of-means estimator pooled over all $20$ clips
($\mathrm{AS} - \text{chained low-band}$ $+0.354$ [$+0.294,
+0.422$]; $\mathrm{Hybrid} - \mathrm{AS}$ $+0.038$ [$+0.019,
+0.060]$).

\begin{table*}[t]
\centering
\small
\begin{tabular}{lcccc}
\toprule
 & \multicolumn{2}{c}{ukiyo\_e retention} & \multicolumn{2}{c}{van\_gogh} \\
\cmidrule(lr){2-3}\cmidrule(lr){4-5}
Comparison ($K{=}6$) & new $10$ clips & pooled $20$ clips & per-frame ($20$) & ratio-of-means ($20$, expl.) \\
\midrule
$\mathrm{AS} - \text{chained low-band}$ & $+0.204$ [$+0.132$, $+0.277$] & $+0.189$ [$+0.150$, $+0.232$] & unstable & $+0.354$ [$+0.294$, $+0.422$] \\
$\mathrm{Hybrid} - \mathrm{AS}$ & $-0.001$ [$-0.031$, $+0.029$] & $+0.001$ [$-0.017$, $+0.019$] & unstable & $+0.038$ [$+0.019$, $+0.060$] \\
\bottomrule
\end{tabular}
\caption{Expanded replication on $10$ fresh seed-$2028$ clips and
pooled across all $20$ clips ($K{=}6$). Per-frame van\_gogh retention
is unstable on the new set due to a $-0.006$ teacher gap on
\emph{upside-down}; the exploratory ratio-of-means column aggregates
numerators and denominators before dividing.}
\label{tab:supp-new10}
\end{table*}

On the same $10$ seed-$2028$ clips (which played no part in
selecting $F$), the two-anchor crossfade control reaches boundary
T\_LPIPS $0.1483$, against $0.1084$ for the $F{=}3$ feather whose
held-out confirmation the main paper reports.

\section{Second teacher (SD-Turbo)}
\label{sec:supp-teacher2}

\Cref{tab:supp-teacher2} gives the full strength sweep for the
SD-Turbo second teacher (calibration and the documented deviation
are in the main paper). At $t_{\mathrm{index}}{=}35$ the teacher's
CSD gain over the source is $0.037$ (ukiyo\_e) and $-0.011$
(van\_gogh), so retention ratios are uninformative there; the
non-ratio metrics replicate the pattern (DINO-to-source AS
$0.870$ vs.\ chained low-band $0.930$ and Hybrid $0.877$; total
T\_LPIPS AS $0.137$, Hybrid $0.135$, chained low-band $0.110$).

\begin{table*}[t]
\centering
\small
\resizebox{\textwidth}{!}{%
\begin{tabular}{lcccc}
\toprule
 & \multicolumn{2}{c}{$\mathrm{AS} - \text{chained low-band}$} & \multicolumn{2}{c}{$\mathrm{Hybrid} - \mathrm{AS}$} \\
\cmidrule(lr){2-3}\cmidrule(lr){4-5}
Strength ($K{=}6$) & ukiyo\_e & van\_gogh & ukiyo\_e & van\_gogh \\
\midrule
$t_{\mathrm{index}}{=}25$ (primary, raw CSD) & $+0.063$ [$+0.041$, $+0.085$] & $+0.072$ [$+0.047$, $+0.099$] & $-0.004$ [$-0.010$, $+0.001$] & $-0.0004$ [$-0.005$, $+0.004$] \\
$t_{\mathrm{index}}{=}25$ (primary, retention) & $+0.331$ [$+0.278$, $+0.384$] & $+0.301$ [$-0.044$, $+0.664$] & $-0.008$ [$-0.040$, $+0.036$] & $+0.079$ [$-0.022$, $+0.194$] \\
$t_{\mathrm{index}}{=}15$ (retention) & $+0.340$ [$+0.289$, $+0.391$] & $+0.306$ [$+0.230$, $+0.382$] & $-0.040$ [$-0.045$, $-0.034$] & $+0.004$ [$-0.009$, $+0.019$] \\
$t_{\mathrm{index}}{=}35$ & \multicolumn{4}{l}{retention uninformative (teacher CSD gain $0.037$ / $-0.011$); see text} \\
\bottomrule
\end{tabular}}
\caption{Second teacher (SD-Turbo, prompt-only img2img) on the
confirmatory set ($K{=}6$). The anchor-over-chain ordering holds at
all three tested strengths ($t_{\mathrm{index}} \in \{15, 25, 35\}$).}
\label{tab:supp-teacher2}
\end{table*}

\section{CSD-independent style metrics}
\label{sec:supp-clip}

\Cref{tab:supp-clip} evaluates all four styles on the confirmatory set
($K{=}6$) using OpenCLIP ViT-B-32 image--image cosine to the style
reference, which shares no weights with CSD. AS outperforms the
chained low-band estimate on all four styles, including watercolor and
ink\_sketch where CSD fails to separate teacher from source, while
IW and FastBlend score higher on style cosine. A VGG19 Gram-matrix
metric separates the teacher from the source only on watercolor
($100\%$ of frames vs.\ $49\%$--$58\%$ on the other styles), where
$\mathrm{AS} - \text{chained low-band}$ is $+0.96{\times}10^{-6}$
$[-1.9{\times}10^{-6}, +4.0{\times}10^{-6}]$ (a tie).

\begin{table*}[t]
\centering
\small
\begin{tabular}{lccccc}
\toprule
Style & $\mathrm{AS} - \text{chained}$ (raw) & AS ret. & chained ret. & $\mathrm{AS} - \mathrm{IW}$ (raw) & $\mathrm{AS} - \text{FastBlend}$ (raw) \\
\midrule
ukiyo\_e    & $+0.031$ [$+0.023$, $+0.042$] & $0.535$ & $0.332$ & $-0.023$ & $-0.054$ \\
van\_gogh   & $+0.057$ [$+0.040$, $+0.074$] & $0.563$ & $0.269$ & $-0.029$ & $-0.066$ \\
watercolor  & $+0.049$ [$+0.039$, $+0.061$] & $0.492$ & $0.249$ & $-0.060$ & $-0.096$ \\
ink\_sketch & $+0.027$ [$+0.017$, $+0.038$] & $0.390$ & $0.203$ & $-0.032$ & $-0.060$ \\
\bottomrule
\end{tabular}
\caption{OpenCLIP ViT-B-32 image--image style cosine on the
confirmatory set ($K{=}6$, all four styles): paired raw differences
and CLIP retention.}
\label{tab:supp-clip}
\end{table*}

\section{Frequency-energy controls}
\label{sec:supp-energy}

The factorial localizes AS's advantage to the high band's transport.
A natural confound is amplitude: iterated warping may simply decay
the high band's energy. With the paper's split ($\sigma = 8$), the
chained estimate's high-band residual RMS halves by anchor age $5$,
while AS holds it roughly constant and the teacher is flat; the low
band stays flat ($0.067$--$0.074$) for every method.

\begin{table*}[t]
\centering
\small
\begin{tabular}{lcccccc}
\toprule
High-band residual RMS & age 0 & 1 & 2 & 3 & 4 & 5 \\
\midrule
Teacher (every frame) & 0.0717 & 0.0714 & 0.0715 & 0.0718 & 0.0718 & 0.0715 \\
Anchor sampling & 0.0507 & 0.0617 & 0.0583 & 0.0552 & 0.0522 & 0.0496 \\
Hybrid & 0.0507 & 0.0605 & 0.0571 & 0.0540 & 0.0510 & 0.0484 \\
Chained low-band & 0.0507 & 0.0401 & 0.0346 & 0.0306 & 0.0276 & 0.0251 \\
\bottomrule
\end{tabular}
\caption{High-band residual RMS by anchor age on the confirmatory
set ($K{=}6$, mean over clips and both styles). Age-$0$ frames use
the keyframe blend (main paper, Sec.~3), so their energy sits below
the teacher's except at $t = 0$.}
\label{tab:supp-energy}
\end{table*}

The amplitude-matched chain control (per-frame, per-channel
rescaling of the chained estimate's high band to AS's high-band
RMS; an oracle diagnostic, not a deployable method) reaches
retention $0.411$ (ukiyo\_e) / $0.414$ (van\_gogh), between the
chained estimate ($0.332$ / $0.174$) and AS ($0.506$ / $0.444$);
the shares of the gap it closes are reported in the main paper
and tabulated in \cref{tab:supp-decomp}. On the saved 8-bit frames
the amplitude-matched chain reaches $95\%$ of AS's high-band RMS
(clamping), so the ukiyo\_e remainder is a slight overestimate.
Without any transport, the teacher's high band alone carries far
more style than its low band alone (retention $0.450$ vs.\
$0.167$ on ukiyo\_e; paired $+0.283$ $[+0.180, +0.422]$;
van\_gogh $+0.169$ [$-0.123$, $+0.598$]). DINO-to-source: AS
$0.829$, Hybrid $0.845$, chained low-band $0.913$,
amplitude-matched chain $0.856$, low-band-only $0.945$,
high-band-only $0.754$.

\section{EbSynth baseline, composed-mask IW, and additional timings}
\label{sec:supp-ebsynth}

\Cref{tab:supp-ebsynth} reports EbSynth~\cite{jamriska2019stylizingvideobyexample}
(official source build, run on the same frozen teacher keyframes at
$K{=}6$ on the confirmatory set). At full strength
($\alpha{=}1$), causal streaming-compatible EbSynth
($\text{EbSynth}_{\text{causal}}$) achieves higher style retention
than AS ($0.899$ vs.\ $0.507$ on ukiyo\_e) at a large content cost
(DINO-to-source $0.628$ vs.\ $0.829$; paired $-0.202$
[$-0.288, -0.129$]) and larger boundary jump ($0.3161$ vs.\ $0.2275$;
paired $+0.089$ $[+0.075, +0.102]$). Offline two-keyframe EbSynth
($\text{EbSynth}_{\text{blend}}$) matches FastBlend's operating point
(boundary $0.1093$, total T\_LPIPS $0.0963$, DINO $0.660$). The
controlled timing run behind the main paper's cost table used the
same binary and invocation as the quality runs ($n = 20$
sequential calls; $40$ frames at $4$ workers) and measured peak
memory of $29$\,MiB per call. At matched content the boundary
ordering reverses (\cref{tab:supp-matched-ebsynth}).

\begin{table*}[t]
\centering
\small
\begin{tabular}{lccccc}
\toprule
Method ($K{=}6$) & ukiyo\_e ret. & van\_gogh raw gain & total T\_LPIPS & boundary T\_LPIPS & DINO-to-source \\
\midrule
AS (reference) & $0.507$ & $+0.088$ & $0.1415$ & $0.2275$ & $0.829$ \\
$\text{EbSynth}_{\text{causal}}$ & $0.899$ & $+0.216$ & $0.1271$ & $0.3161$ & $0.628$ \\
$\text{EbSynth}_{\text{blend}}$ (offline) & $0.887$ & $+0.195$ & $0.0963$ & $0.1093$ & $0.660$ \\
\bottomrule
\end{tabular}
\caption{EbSynth baseline on the confirmatory set ($K{=}6$, same
frozen teacher keyframes).}
\label{tab:supp-ebsynth}
\end{table*}

\paragraph{Composed-mask image warp.}
Stored IW differs from AS in three factors besides the transported
quantity: it warps with the direct $K{=}6$ RAFT flow instead of the
composed adjacent flow, masks with the direct-flow confidence
instead of AS's composed mask, and shows raw teacher frames at
keyframes instead of AS's blended keyframe output (mean keyframe
difference $5.70/255$). We re-ran image warp on the confirmatory
set ($K{=}6$) changing one factor at a time. Changing only the mask
(composed mask, chained low-band fallback) leaves style,
LPIPS-to-teacher, and DINO almost unchanged and mainly lowers
within-interval flicker (internal T\_LPIPS $-0.014$); switching to
the composed flow lowers flicker further (internal $-0.010$);
adopting AS's blended keyframe frames removes $0.076$ of stored IW's
$0.102$ boundary excess and lowers ukiyo\_e retention by $0.033$.
With all three matched, image warp differs from AS only in the
transported quantity: retention $0.624$ (ukiyo\_e) / $0.476$
(van\_gogh), total / internal / boundary T\_LPIPS $0.110$ / $0.085$
/ $0.253$, LPIPS-to-teacher $0.276$, DINO-to-source $0.760$. Paired
against AS: retention $+0.118$ [$+0.064$, $+0.194$] (ukiyo\_e;
van\_gogh $+0.032$ [$-0.100$, $+0.217$], n.s.), total T\_LPIPS
$-0.031$ [$-0.038$, $-0.023$], boundary $+0.025$ [$+0.017$,
$+0.035$], LPIPS-to-teacher $-0.023$ [$-0.035$, $-0.012$], DINO
$-0.070$ [$-0.093$, $-0.047$]. Every sign of the stored-IW
comparison survives the matching. (The watercolor / ink\_sketch IW
columns of \cref{tab:supp-clip} still use the direct mask.)

\paragraph{Additional $K{=}9$ and same-GPU timings.}
In a repeat RTX 5080 Laptop timing session under the same
protocol as the main paper's comparison (main paper, Sec.~4),
$K{=}9$ amortized runtimes are $42.1$\,ms/frame (Hybrid),
$41.0$\,ms/frame (AS), and $40.3$\,ms/frame (chained low-band),
agreeing with the session behind the main paper's Tab.~1
($42.02$ / $40.83$ / $40.25$\,ms/frame) to within
$0.2$\,ms/frame (repeat SD $\leq 0.31$\,ms/frame), with
non-keyframe per-frame costs of $21.1$, $19.8$, and $19.1$\,ms/frame,
respectively (propagation-only estimates: IW $17.5$\,ms/frame,
chained image warp $16.6$\,ms/frame). On an RTX 4090 ($6$ development
clips $\times$ $2$ styles, $K{=}6$), the per-frame teacher costs
$161.1$\,ms/frame; AS's end-to-end cost splits into $23.0$\,ms
propagation-only and a $163.2$\,ms keyframe; FastBlend costs
$216.1$\,ms/frame propagation-only including $\approx 61.6$\,ms PNG
I/O ($\approx 154.5$\,ms compute), or $\approx 181$\,ms/frame
end-to-end without PNG I/O.

\section{Metric robustness, mechanism decomposition, and confirmatory p-values}
\label{sec:supp-r5}

This section provides the full exploratory tables and confirmatory
permutation $p$-values accompanying the mechanism analyses on the
confirmatory set ($K{=}6$, $n = 10$ clips).

\paragraph{Unnormalized CSD gain and ratio-of-means factorial.}
\Cref{tab:supp-raw-csd} reports the factorial in unnormalized raw CSD
gain ($\mathrm{method} - \mathrm{source}$) so that no per-frame
division enters. Under a ratio-of-means estimator, retention
levels are $0.511$ (Hybrid), $0.508$ (AS), and $0.332$ (chained
low-band) on ukiyo\_e, and $0.720$, $0.691$, and $0.334$ on
van\_gogh; paired $\mathrm{Hybrid} - \mathrm{AS}$ is $+0.003$
$[-0.012, +0.020]$ (ukiyo\_e) and $+0.029$ $[+0.007, +0.059]$
(van\_gogh), while $\mathrm{AS} - \text{chained low-band}$ is
$+0.176$ $[+0.147, +0.206]$ and $+0.356$ $[+0.304, +0.436]$.

\begin{table*}[t]
\centering
\small
\begin{tabular}{lcc}
\toprule
Quantity ($K{=}6$, confirmatory set) & ukiyo\_e & van\_gogh \\
\midrule
Teacher (every frame) & $+0.277$ & $+0.225$ \\
Anchor, anchor (AS) & $+0.141$ & $+0.156$ \\
Chain, anchor (Hybrid) & $+0.141$ & $+0.162$ \\
Anchor, chain & $+0.094$ & $+0.074$ \\
Chain, chain (chained low-band) & $+0.092$ & $+0.075$ \\
\midrule
$\mathrm{AS} - \text{chained low-band}$ & $+0.049$ [$+0.039$, $+0.060$] & $+0.080$ [$+0.064$, $+0.097$] \\
High-band anchor main effect & $+0.048$ [$+0.039$, $+0.059$] & $+0.084$ [$+0.069$, $+0.100$] \\
Low-band anchor main effect & $+0.001$ [$-0.003$, $+0.005$] & $-0.004$ [$-0.008$, $+0.001$] \\
Band $\times$ transport interaction & $-0.003$ & $-0.005$ \\
$\mathrm{Hybrid} - \mathrm{AS}$ & $+0.001$ [$-0.003$, $+0.005$] & $+0.007$ [$+0.002$, $+0.011$] \\
\bottomrule
\end{tabular}
\caption{Unnormalized raw CSD style gain ($\mathrm{method} -
\mathrm{source}$) on the confirmatory set ($K{=}6$, exploratory).}
\label{tab:supp-raw-csd}
\end{table*}

\paragraph{CLIP $\mathrm{Hybrid} - \mathrm{AS}$ comparison and denominator audit.}
\Cref{tab:supp-clip-hyb} shows that on the CSD-independent CLIP
cosine metric, $\mathrm{Hybrid} - \mathrm{AS}$ is negative in all
four styles ($10/10$ clips each). \Cref{tab:supp-denom} audits the
per-clip and per-frame CSD retention denominators
($\mathrm{teacher} - \mathrm{source}$): no confirmatory-set clip has
a mean gap below $0.05$, while van\_gogh has $6$ of $110$ scored
frames below $0.05$ and $2$ frames $\le 0$, explaining its wider
per-frame retention intervals.

\begin{table*}[t]
\centering
\small
\begin{tabular}{lccc}
\toprule
Style & $\mathrm{Hybrid} - \mathrm{AS}$ raw CLIP cosine & CLIP retention diff. & Clips $< 0$ \\
\midrule
ukiyo\_e    & $-0.007$ [$-0.009$, $-0.005$] & $-0.046$ & $10/10$ \\
van\_gogh   & $-0.014$ [$-0.021$, $-0.008$] & $-0.075$ & $10/10$ \\
watercolor  & $-0.016$ [$-0.021$, $-0.010$] & $-0.081$ & $10/10$ \\
ink\_sketch & $-0.009$ [$-0.013$, $-0.005$] & $-0.060$ & $10/10$ \\
\bottomrule
\end{tabular}
\caption{Paired $\mathrm{Hybrid} - \mathrm{AS}$ on the
CSD-independent OpenCLIP ViT-B-32 cosine metric ($K{=}6$,
confirmatory set, exploratory).}
\label{tab:supp-clip-hyb}
\end{table*}

\begin{table*}[t]
\centering
\small
\begin{tabular}{lccccc}
\toprule
Style & Clip min / median / max & Clips $< 0.05$ & Frames $< 0.05$ & Frames $\le 0$ \\
\midrule
ukiyo\_e  & $0.203$ / $0.264$ / $0.336$ & $0/10$ & $0/110$ & $0/110$ \\
van\_gogh & $0.054$ / $0.216$ / $0.472$ & $0/10$ & $6/110$ & $2/110$ \\
\bottomrule
\end{tabular}
\caption{Audit of the CSD retention denominator ($\mathrm{teacher} -
\mathrm{source}$) on the confirmatory set ($10$ clips, $110$ scored
frames).}
\label{tab:supp-denom}
\end{table*}

\paragraph{Mechanism decomposition and confirmatory permutation $p$-values.}
\Cref{tab:supp-decomp} summarizes the two-part decomposition of the
$\mathrm{AS} - \text{chained low-band}$ retention gap into
high-frequency amplitude preservation and an amplitude-unexplained remainder.
\Cref{tab:supp-pvals} lists two-sided paired sign-flip permutation
$p$-values ($10{,}000$ flips; exact enumeration over all $2^{10}$
flips agrees) alongside Holm-adjusted $p$-values across the
confirmatory tests; these $p$-values test difference from zero and do
not establish equivalence. Finally, \cref{tab:supp-cell-ci} provides
per-cell 95\% bootstrap CIs for the main confirmatory tables (note
that van\_gogh per-cell intervals are wide due to the per-frame
denominator above, whereas paired differences are the primary
inferential objects).

\paragraph{Equivalence of Hybrid and AS (exploratory).}
Because the pre-registered decision rule can only fail to
support Hybrid, we add a post-hoc equivalence check (two
one-sided tests at the $\pm 0.05$ retention margin the rule
used). On the ratio-of-means estimator, Hybrid $-$ AS is
$+0.003$ [$-0.012$, $+0.020$] on ukiyo\_e, inside the margin:
the two are equivalent there. On van\_gogh it is $+0.029$
[$+0.007$, $+0.059$], which the margin does not contain, so
equivalence is not established; the point estimate favours
Hybrid slightly. The per-frame estimator leads to the same
conclusion on both styles. The supported statement is
therefore narrow: splitting the bands buys nothing measurable
on ukiyo\_e, and at most a small, not clearly bounded gain on
van\_gogh.

\begin{table*}[t]
\centering
\small
\begin{tabular}{lcccc}
\toprule
Style & Total gap ($\mathrm{AS} - \text{chained}$) & Closed by amplitude & Share closed & Unexplained remainder \\
\midrule
ukiyo\_e  & $+0.174$ & $+0.079$ [$+0.044$, $+0.114$] & $45.4\%$ [$30\%$, $60\%$] & $+0.095$ [$+0.075$, $+0.114$] ($54.6\%$) \\
van\_gogh & $+0.270$ & $+0.240$ [$+0.087$, $+0.351$] & $89.1\%$ & $+0.029$ [$-0.069$, $+0.098$] (n.s.) \\
\bottomrule
\end{tabular}
\caption{Decomposition of the $\mathrm{AS} - \text{chained low-band}$
CSD retention gap via the high-frequency amplitude-matched chain
control ($K{=}6$, confirmatory set, exploratory).}
\label{tab:supp-decomp}
\end{table*}

\begin{table*}[t]
\centering
\small
\resizebox{\textwidth}{!}{%
\begin{tabular}{llcc}
\toprule
Comparison & Metric / Style & Raw $p$ & Holm-adjusted $p$ \\
\midrule
$\mathrm{Hybrid} - \mathrm{AS}$ & CSD retention (ukiyo\_e) & $0.77$ & $1.000$ \\
$\mathrm{Hybrid} - \mathrm{AS}$ & CSD retention (van\_gogh; raw gain $+0.007$ [$+0.002, +0.011$]) & $0.17$ & $0.687$ \\
$\mathrm{Hybrid} - (\mathrm{AS}+\mathrm{EMA})$ & CSD retention (ukiyo\_e) & $0.002$ & $0.018$ \\
$\mathrm{Hybrid} - (\mathrm{AS}+\mathrm{EMA})$ & CSD retention (van\_gogh) & $0.003$ & $0.020$ \\
High-band anchor main effect & CSD retention (ukiyo\_e) & $0.002$ & $0.018$ \\
High-band anchor main effect & CSD retention (van\_gogh; raw gain $+0.084$ [$+0.069, +0.100$]) & $0.065$ & $0.326$ \\
Low-band anchor main effect & CSD retention (ukiyo\_e) & $0.78$ & $1.000$ \\
Low-band anchor main effect & CSD retention (van\_gogh) & $0.44$ & $1.000$ \\
Pooled $\mathrm{AS} - \text{chained low-band}$ ($20$ clips) & CSD retention (ukiyo\_e) & $<0.001$ & $0.001$ \\
Feather confirmation ($10$ fresh clips) & Boundary T\_LPIPS & $0.002$ & $0.018$ \\
\bottomrule
\end{tabular}}
\caption{Two-sided paired sign-flip permutation $p$-values ($10{,}000$
flips) and Holm-adjusted $p$-values for the confirmatory
comparisons. The factorial effects, Hybrid $-$ AS and the pooled
AS $-$ chained low-band rows are the pre-registered family; the
feather row is a later confirmation run on fresh clips, and the
AS+EMA rows are exploratory. The Holm adjustment is applied
across all ten rows shown, including the exploratory ones,
which is conservative for the pre-registered family.}
\label{tab:supp-pvals}
\end{table*}

\begin{table*}[t]
\centering
\small
\begin{tabular}{lcc}
\toprule
Method / Cell ($K{=}6$, confirmatory set) & ukiyo\_e retention [95\% CI] & van\_gogh retention [95\% CI] \\
\midrule
Chain, chain (chained low-band) & $0.332$ [$0.228$, $0.453$] & $0.174$ [$-0.114$, $0.371$] \\
Anchor, chain & $0.339$ [$0.235$, $0.459$] & $0.172$ [$-0.081$, $0.356$] \\
Chain, anchor (Hybrid) & $0.509$ [$0.399$, $0.631$] & $0.472$ [$-0.067$, $0.815$] \\
Anchor, anchor (AS) & $0.506$ [$0.406$, $0.615$] & $0.444$ [$-0.068$, $0.764$] \\
AS + EMA (exploratory) & $0.479$ [$0.375$, $0.588$] & $0.387$ [$-0.155$, $0.724$] \\
\midrule
Image warp (IW, stored) & $0.659$ [$0.548$, $0.772$] & $0.618$ [$0.411$, $0.787$] \\
Image warp (matched flow/mask/keyframes) & $0.624$ [$0.498$, $0.754$] & $0.476$ [$0.118$, $0.744$] \\
AS + feather ($F{=}3$) & $0.618$ [$0.527$, $0.710$] & $0.597$ [$0.311$, $0.795$] \\
\bottomrule
\end{tabular}
\caption{Per-cell 95\% clip-cluster bootstrap confidence intervals on
the confirmatory set ($K{=}6$, $n = 10$ clips). Van\_gogh per-cell
intervals are wide due to two frames with $\le 0$ denominator; paired
differences are the primary inferential quantities.}
\label{tab:supp-cell-ci}
\end{table*}

\section{Matched operating points}
\label{sec:supp-matched}

Operating points are compared at matched content preservation on
the confirmatory set ($K{=}6$) by blending each method's output
toward the source frame, $\mathrm{output} = \mathrm{source} +
\alpha(\mathrm{method} - \mathrm{source})$, and recomputing all
metrics; the FastBlend matches are reported in the main paper.
Completing them: scaling matched image warp to FastBlend's
content level ($\alpha = 1.83$) likewise leaves it behind
FastBlend on retention ($-0.439$ ukiyo\_e / $-0.112$ van\_gogh)
and flicker ($+0.062$ total T\_LPIPS; $+0.223$ boundary).
\Cref{tab:supp-matched-ebsynth} gives the corresponding
content-matched causal EbSynth comparison.

\begin{table*}[t]
\centering
\small
\resizebox{\textwidth}{!}{%
\begin{tabular}{lccccccc}
\toprule
Method / operating point ($K{=}6$) & $\alpha$ & ukiyo\_e ret. & van\_gogh ret. & DINO-to-source & total / internal / boundary T\_LPIPS & LPIPS-to-teacher \\
\midrule
AS (reference) & $1.00$ & $0.506$ & $0.444$ & $0.829$ & $0.142$ / $0.126$ / $0.228$ & $0.267$ \\
Causal EbSynth at AS content & $0.68$ & \multicolumn{2}{c}{pooled diff.\ $+0.001$ [$-0.178$, $+0.272$]} & $0.831$ & $0.099$ / -- / $0.219$ & $0.223$ \\
\bottomrule
\end{tabular}}
\caption{Content-matched causal EbSynth comparison on the
confirmatory set ($K{=}6$, $n = 10$ clips). At $\alpha = 0.68$,
pooled CSD retention is not separated from AS ($+0.001$
$[-0.178, +0.272]$; per style $+0.020$ [$-0.041, +0.075$] on
ukiyo\_e and $-0.018$ [$-0.327, +0.485$] on van\_gogh), while AS keeps a higher raw CSD gain on
van\_gogh by $+0.056$), and EbSynth has lower T\_LPIPS and
LPIPS-to-teacher.}
\label{tab:supp-matched-ebsynth}
\end{table*}

\section{Second content-preserving teacher}
\label{sec:supp-teacher-cp}

\Cref{tab:supp-teacher-cp-dev} gives the development-set
calibration behind the main paper's second content-preserving
teacher: strengths $0.40$ and $0.45$ produce identical frames
under four inference steps, so strength $0.50$ (teacher CSD gain
$0.191$, DINO-to-source $0.685$ on development clips) was
selected. \Cref{tab:supp-teacher-cp} reports the full
confirmatory-set results ($K{=}6$).

What the strength value varies at four steps is the number of
denoising steps: the pipeline keeps $\lfloor 4 \times
\mathrm{strength} \rfloor$ steps, so $0.40$ and $0.45$ run one
step (timestep $249$) and $0.50$ through $0.70$ run two steps
(timesteps $499$, $249$) from the same initial noise level. We
verified this at the frame level on a development clip:
strengths $0.50$, $0.55$, $0.60$ and $0.65$ render identical
frames at ControlNet scale $1.0$, and $0.50$ against $0.65$
renders identical frames at scale $0.8$. The second teacher
therefore differs from the primary teacher, at the level of the
frames produced, in ControlNet scale alone ($1.0$ against
$0.8$); both run two denoising steps. The calibration above is
a choice between a one-step teacher ($0.40$) and a two-step
teacher ($0.50$) at the stronger ControlNet scale.

\begin{table*}[t]
\centering
\small
\begin{tabular}{lcc}
\toprule
Teacher configuration (development set) & Teacher CSD gain & Teacher DINO-to-source \\
\midrule
Primary teacher (ControlNet $0.8$, strength $0.65$) & $0.248$ & $0.660$ \\
ControlNet $1.0$, strength $0.40$ (identical to $0.45$ at $4$ steps) & $0.039$ & $0.853$ \\
ControlNet $1.0$, strength $0.50$ (selected) & $0.191$ & $0.685$ \\
\bottomrule
\end{tabular}
\caption{Development-set calibration of the second content-preserving
teacher (ControlNet scale $1.0$, four inference steps).}
\label{tab:supp-teacher-cp-dev}
\end{table*}

\begin{table*}[t]
\centering
\small
\begin{tabular}{lccccc}
\toprule
Method ($K{=}6$, ControlNet $1.0$, strength $0.50$) & ukiyo\_e ret. & van\_gogh ret. & DINO-to-source & total T\_LPIPS & boundary T\_LPIPS \\
\midrule
Teacher (every frame) & $1.000$ & $1.000$ & $0.675$ & $0.243$ & $0.243$ \\
Chained low-band baseline & $0.361$ & $0.321$ & $0.921$ & $0.111$ & $0.172$ \\
Anchor sampling (AS) & $0.530$ & $0.716$ & $0.843$ & $0.141$ & $0.220$ \\
Hybrid & $0.535$ & $0.794$ & $0.857$ & $0.138$ & $0.215$ \\
\bottomrule
\end{tabular}
\caption{Confirmatory-set results ($K{=}6$, $n = 10$ clips) under
the second content-preserving teacher.}
\label{tab:supp-teacher-cp}
\end{table*}

\section{A feed-forward teacher}
\label{sec:supp-sanet}

The two diffusion teachers above share a model family, and the
second differs from the primary only in ControlNet scale
(\cref{sec:supp-teacher-cp}). This section changes the teacher
family itself: SANet, a feed-forward arbitrary style transfer
network (official release weights, style strength $1.0$), is
run on every frame of the confirmatory set with the same two
style reference images, and anchor sampling and the chained
low-band baseline are run over its residuals with the identical
flow, confidence and keyframe machinery ($K{=}6$). The teacher
is fast ($25.7$ / $26.8$ ms/frame in half precision on the
RTX 5080 Laptop, ukiyo\_e / van\_gogh) and its content
preservation is low (DINO-to-source $0.251$ / $0.409$), so this
is a test of the transport claim, not a deployment
configuration.

\begin{table*}[t]
\centering
\small
\begin{tabular}{lccccc}
\toprule
Method ($K{=}6$, SANet teacher) & \multicolumn{2}{c}{Raw gain} & \multicolumn{2}{c}{Retention} & DINO \\
 & ukiyo\_e & van\_gogh & ukiyo\_e & van\_gogh & \\
\midrule
Teacher (every frame) & $0.366$ & $0.339$ & $1.000$ & $1.000$ & $0.330$ \\
Anchor sampling & $0.235$ & $0.225$ & $0.652$ & $0.669$ & $0.544$ \\
Hybrid & $0.234$ & $0.218$ & $0.652$ & $0.647$ & $0.561$ \\
Chained low-band & $0.182$ & $0.140$ & $0.506$ & $0.410$ & $0.712$ \\
\bottomrule
\end{tabular}
\caption{Anchor over chain replicates under a feed-forward
teacher (confirmatory set, $n = 10$ clips). AS exceeds the
chained low-band baseline by $+0.053$ [$+0.044$, $+0.063$]
(ukiyo\_e) and $+0.085$ [$+0.060$, $+0.114$] (van\_gogh) in raw
CSD gain, paired by clip; Hybrid again matches AS. DINO is the
mean over the two styles. The teacher's gain denominators are
well separated from zero here, and the per-frame and
ratio-of-means retention estimators agree to within $0.01$.}
\label{tab:supp-sanet}
\end{table*}

The ordering the paper rests on is therefore not an artefact
of the diffusion teacher: with a feed-forward teacher whose
residual has a different structure (no denoising, no text or
ControlNet conditioning), anchoring the residual at keyframes
still keeps more of the teacher's style gain than chaining the
low band, by margins comparable to the diffusion case.

\section{Transport vehicle and boundary feathering}
\label{sec:supp-vehicle}

What is transported matters as much as how. The vehicle
comparison itself is in \cref{sec:supp-ebsynth} (the composed-mask
image-warp control) and the temporal record is in
\cref{tab:supp-temporal}: under the same flow, mask and keyframe
construction, warping the stylized image keeps at least as much
CSD style as warping the residual (retention $+0.118$
[$+0.064$, $+0.194$] on ukiyo\_e, not separated on van\_gogh)
and sits closer to the teacher, while the residual vehicle
keeps the smaller boundary jump (image warp's is larger by
$0.025$ [$+0.017$, $+0.035$]) and the higher content
similarity (image warp's DINO is lower by $0.070$
[$-0.093$, $-0.047$]). The choice of
vehicle is therefore a trade-off, not a style win for the
residual; lookahead feathering reduces the boundary jump for
both vehicles while costing more content similarity on the
image vehicle. This section is the pointer to that record; the
feathering failure mode that the averages hide is shown next,
in \cref{sec:supp-feather-fail}.

\begin{table*}[tb]
\centering
\scriptsize
\begin{tabular}{lcccccc}
\toprule
 & \multicolumn{2}{c}{CSD retention} & \multicolumn{3}{c}{T\_LPIPS} & \\
\cmidrule(lr){2-3}\cmidrule(lr){4-6}
Method & ukiyo\_e & van\_gogh & total & internal & boundary & DINO-to-source \\
\midrule
AS        & 0.51 & 0.44 & 0.142 & 0.126 & 0.228 & \textbf{0.829} \\
IW        & 0.66 & 0.62 & 0.126 & 0.090 & 0.329 & 0.747 \\
IW (matched) & 0.62 & 0.48 & \underline{0.110} & \textbf{0.085} & 0.253 & 0.760 \\
Chained image warp & \underline{0.73} & \underline{0.63} & 0.251 & 0.203 & 0.514 & -- \\
FastBlend (external) & \textbf{0.84} & \textbf{0.72} & \textbf{0.093} & \underline{0.088} & 0.119 & 0.690 \\
AS + feather ($F{=}1$)     & 0.55 & 0.42 & 0.133 & 0.137 & \underline{0.110} & -- \\
AS + feather ($F{=}2$)     & 0.59 & 0.45 & 0.129 & 0.134 & \textbf{0.101} & -- \\
AS + feather ($F{=}3$)     & 0.62 & 0.60 & 0.124 & 0.127 & \underline{0.110} & \underline{0.781} \\
\bottomrule
\end{tabular}
\caption{Transport vehicle, propagation structure, and boundary feathering on the confirmatory set
($K{=}6$). Stored IW uses the direct $K{=}6$ RAFT flow and mask; the matched row also matches AS's flow, mask, and keyframe frames (supplementary). Feather rows are AS with an $F$-frame lookahead blend toward the next keyframe (\cref{sec:feather}). Bold marks the best value in each column and underline the
second best.}
\label{tab:vehicle}
\end{table*}

\section{Feather failure mode}
\label{sec:supp-feather-fail}

\begin{figure*}[t]
\centering
\includegraphics[width=\textwidth]{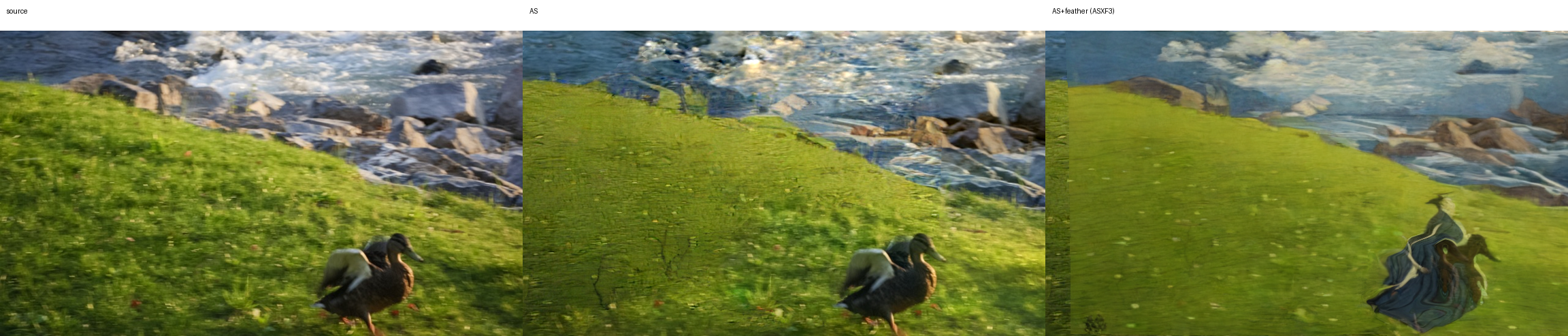}
\caption{Feather failure mode on \emph{mallard-fly} (frame $35$,
immediately before the keyframe boundary at frame $36$). Because the
teacher replaces the flying mallard at keyframe $36$, lookahead
feathering imports that subject replacement up to three frames
early, whereas anchor sampling preserves the original bird.}
\label{fig:supp-feather-fail}
\end{figure*}

While lookahead feathering reduces boundary jumps on average, blending
the next keyframe in early can import content errors from that
keyframe prior to the boundary. \Cref{fig:supp-feather-fail} shows
the worst observed case on \emph{mallard-fly} at frame $35$ (one
frame before the keyframe boundary at frame $36$): the teacher
replaces the subject at frame $36$, and feathering introduces that
replacement up to three frames early, whereas anchor sampling keeps
the original subject.
\section{Qualitative examples}
\label{sec:supp-qual}

\begin{figure*}[t]
\centering
\includegraphics[width=\textwidth]{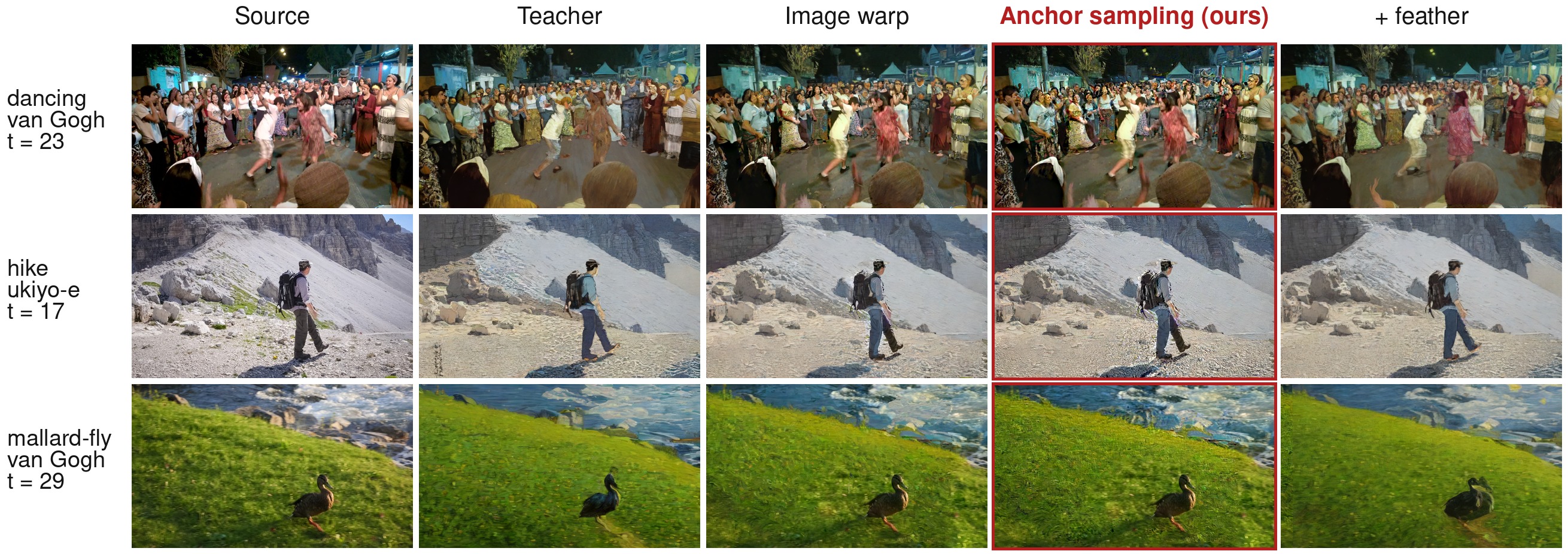}
\caption{Qualitative comparison at anchor age five ($K{=}6$).
Columns: source, teacher, image warp (IW), anchor sampling (AS),
AS + feather. Rows: \emph{dancing} (van Gogh), \emph{hike} (ukiyo-e),
\emph{mallard-fly} (van Gogh).}
\label{fig:supp-qual}
\end{figure*}

\Cref{fig:supp-qual} shows one frame at anchor age five for
three clips, with the source, the teacher, image warp, anchor
sampling and anchor sampling with feathering side by side. The
clips were chosen to span the two main styles and a range of
motion (a slow dance, a walking shot, and a fast flying
subject); the frame is the last frame before a keyframe refresh,
where reuse error is largest.

\section{Implementation checks}
\label{sec:supp-audit}

Three checks bound how far the numbers in the main paper can be
trusted. (i)~The factorial implementation is exact: the chain/anchor
and anchor/anchor cells reproduce Hybrid and AS bit-for-bit on all
$40$ cells (mean absolute difference $0.0$). (ii)~An independent rebuild of AS from the stored frames matches the original outputs at a mean of $0.25/255$ ($99.96\%$ of pixels within $2/255$). (iii)~A cross-host audit of the stored frames against the online teacher (mean and p99 $\le 2/255$) passed for all clips except cows and could not run on blackswan (frames absent on the second host); all quality numbers come from a single set of stored frames on one host,
while timing (main paper, Sec.~4) is measured on the second.

\section{Per-band refresh rate and compute}
\label{sec:supp-bandrefresh}

The factorial results separate the two bands by where they are
sampled. A complementary question is how often each band must be
recomputed. On the confirmatory set ($10$ clips, both styles,
$60$ frames), we rebuilt the output with each band either
anchor-warped from its own last refresh or taken fresh from the
teacher frame at that band's refresh frames, with refresh
intervals $K_{low}, K_{high} \in \{6, 3, 1\}$; a band refresh uses
the same keyframe construction as anchor sampling itself. The
$(1,1)$ condition reproduces the teacher frames exactly (mean
absolute difference $0$), and $(6,6)$ reproduces the stored anchor
sampling frames (mean absolute difference below $10^{-6}$ of the
$8$-bit range). A refresh of either band requires a teacher run,
so cost is set by the union of the two bands' refresh frames:
refreshing one band every frame costs the same teacher time as
refreshing both. The ms/frame values in
\cref{tab:supp-bandrefresh} are model estimates built from the
measured components (teacher $222.11$ ms/frame, propagation
$19.83$ ms/frame on the timing host); they sit slightly above the
directly measured pipeline means of the main paper because the
model prices every keyframe at the full teacher cost.

\begin{table*}[t]
\centering
\small
\begin{tabular}{lccccccccc}
\toprule
$(K_{low}, K_{high})$ & \multicolumn{2}{c}{Raw CSD gain} & \multicolumn{2}{c}{Retention} & DINO & T\_LPIPS & LPIPS to & Teacher & ms/ \\
 & ukiyo\_e & van\_gogh & ukiyo\_e & van\_gogh & & total & teacher & fraction & frame \\
\midrule
$(6,6)$ (anchor sampling) & $0.141$ & $0.156$ & $0.506$ & $0.444$ & $0.829$ & $0.142$ & $0.267$ & $1/6$ & $53.5$ \\
$(6,3)$ & $0.166$ & $0.147$ & $0.603$ & $0.803$ & $0.804$ & $0.167$ & $0.225$ & $1/3$ & $90.6$ \\
$(6,1)$ & $0.226$ & $0.166$ & $0.808$ & $0.838$ & $0.678$ & $0.229$ & $0.037$ & $1$ & $238.6$ \\
$(3,6)$ & $0.151$ & $0.167$ & $0.548$ & $0.440$ & $0.829$ & $0.144$ & $0.260$ & $1/3$ & $90.6$ \\
$(3,3)$ & $0.183$ & $0.170$ & $0.668$ & $0.875$ & $0.791$ & $0.172$ & $0.219$ & $1/3$ & $87.3$ \\
$(3,1)$ & $0.248$ & $0.192$ & $0.888$ & $0.816$ & $0.669$ & $0.231$ & $0.028$ & $1$ & $235.3$ \\
$(1,6)$ & $0.168$ & $0.178$ & $0.614$ & $0.450$ & $0.821$ & $0.159$ & $0.241$ & $1$ & $238.6$ \\
$(1,3)$ & $0.202$ & $0.180$ & $0.741$ & $0.903$ & $0.786$ & $0.185$ & $0.197$ & $1$ & $235.3$ \\
$(1,1)$ (teacher) & $0.277$ & $0.225$ & $1.000$ & $1.000$ & $0.638$ & $0.254$ & $0.000$ & $1$ & $222.1$ \\
\bottomrule
\end{tabular}
\caption{Per-band refresh grid on the confirmatory set. Raw CSD
gain and retention are per style; DINO, T\_LPIPS and LPIPS to
teacher are means over the $20$ clip--style units. CSD and DINO
are measured on frames $5, 10, \dots, 55$; T\_LPIPS uses all
adjacent pairs; LPIPS to teacher uses all $60$ frames. Read raw
gain as the primary column: the van\_gogh retention column is
the per-frame ratio estimator, whose per-frame denominators
(the teacher's gain on that frame) are near zero or negative on
a subset of frames, so the column is not monotone in raw gain
--- $(6,3)$ pairs a lower raw gain ($0.147$) with a higher
retention ($0.803$) than anchor sampling ($0.156$, $0.444$)
because a few frames with tiny denominators dominate the mean
ratio. The ukiyo\_e column, whose denominators are well
separated from zero, is monotone. See
\cref{sec:supp-metricreading}.}
\label{tab:supp-bandrefresh}
\end{table*}

At equal teacher cost the two bands buy different things
(\cref{tab:supp-bandrefresh-equal}). Refreshing the high band
every frame moves the output most of the way to the teacher
(LPIPS to teacher $0.267 \to 0.037$) and, on ukiyo\_e, adds about
three times the CSD gain of low-band refresh; it also imports
most of the teacher's frame-to-frame change and content drift
(T\_LPIPS $0.142 \to 0.229$ against the teacher's $0.254$; DINO
$0.829 \to 0.678$ against the teacher's $0.638$). Low-band
refresh adds a smaller gain that is similar on both styles, with
a small temporal cost and no resolved change in DINO. On
van\_gogh the two bands are not ordered by CSD gain at this
sample size (difference $-0.011$ [$-0.040$, $+0.012$]). The same
pattern holds one step down in cost: at teacher fraction $1/3$,
$(6,3)$ adds $+0.026$ [$+0.018$, $+0.034$] raw gain on ukiyo\_e
and $-0.008$ [$-0.017$, $+0.001$] on van\_gogh, while $(3,6)$
adds $+0.011$ [$+0.008$, $+0.013$] and $+0.012$ [$+0.009$,
$+0.015$], with pooled T\_LPIPS changes of $+0.026$ and $+0.002$
respectively.

\begin{table}[t]
\centering
\small
\resizebox{\columnwidth}{!}{%
\begin{tabular}{lccc}
\toprule
Change vs.\ $(6,6)$ & $(6,1)$ & $(1,6)$ & $(6,1) - (1,6)$ \\
\midrule
Raw gain, ukiyo\_e & $+0.086$ [$+0.066$, $+0.106$] & $+0.028$ [$+0.019$, $+0.037$] & $+0.058$ [$+0.033$, $+0.084$] \\
Raw gain, van\_gogh & $+0.011$ [$-0.019$, $+0.038$] & $+0.022$ [$+0.013$, $+0.033$] & $-0.011$ [$-0.040$, $+0.012$] \\
DINO (pooled) & $-0.152$ [$-0.220$, $-0.098$] & $-0.008$ [$-0.017$, $+0.001$] & $-0.144$ [$-0.207$, $-0.094$] \\
T\_LPIPS (pooled) & $+0.087$ [$+0.081$, $+0.094$] & $+0.018$ [$+0.015$, $+0.021$] & $+0.069$ [$+0.062$, $+0.076$] \\
\bottomrule
\end{tabular}}
\caption{Equal-cost band comparison: refreshing one band every
frame against anchor sampling at $K{=}6$. Both conditions run the
teacher on every frame. Intervals are clip-cluster bootstrap
$95\%$ intervals.}
\label{tab:supp-bandrefresh-equal}
\end{table}

\section{Per-band factorial across teachers, lengths and a second domain}
\label{sec:supp-bandfactorial}

The confirmatory factorial attributes the retention gain to the
anchored high band. \Cref{tab:supp-bandfactorial} repeats the
same $2 \times 2$ decomposition (bands chained or anchored) in
six further settings: the full-length sequences (all frames, and
frames $\ge 60$ separately), the second content-preserving
teacher, SD-Turbo at three sampling strengths, and a second
domain. The second domain is Aria Everyday Activities
(egocentric indoor recordings): ten recordings, the first
recording of the two lowest-numbered scripts at each of five
locations, cut to $60$ frames from the midpoint, run with the
frozen teacher at $K{=}6$.

\begin{table*}[t]
\centering
\small
\resizebox{\textwidth}{!}{%
\begin{tabular}{llccc}
\toprule
Setting & Style & High-band main & Low-band main & Interaction \\
\midrule
Full length, all frames & ukiyo\_e & $+0.052$ [$+0.042$, $+0.064$] & $+0.001$ [$-0.003$, $+0.005$] & $-0.004$ [$-0.008$, $-0.001$] \\
 & van\_gogh & $+0.091$ [$+0.072$, $+0.110$] & $-0.004$ [$-0.008$, $+0.001$] & $-0.005$ [$-0.007$, $-0.002$] \\
Full length, frames $\ge 60$ & ukiyo\_e & $+0.058$ [$+0.045$, $+0.073$] & $+0.004$ [$-0.001$, $+0.009$] & $-0.002$ [$-0.006$, $+0.003$] \\
 & van\_gogh & $+0.099$ [$+0.077$, $+0.124$] & $-0.002$ [$-0.007$, $+0.003$] & $-0.005$ [$-0.009$, $+0.001$] \\
Second teacher (ControlNet $1.0$, strength $0.50$) & ukiyo\_e & $+0.038$ [$+0.029$, $+0.047$] & $+0.000$ [$-0.005$, $+0.004$] & $-0.003$ [$-0.007$, $+0.000$] \\
 & van\_gogh & $+0.068$ [$+0.055$, $+0.081$] & $-0.004$ [$-0.009$, $+0.000$] & $-0.004$ [$-0.007$, $-0.001$] \\
SD-Turbo, $t$ index $25$ & ukiyo\_e & $+0.063$ [$+0.043$, $+0.081$] & $+0.001$ [$-0.004$, $+0.006$] & $+0.007$ [$+0.004$, $+0.010$] \\
 & van\_gogh & $+0.073$ [$+0.049$, $+0.098$] & $-0.001$ [$-0.004$, $+0.004$] & $+0.002$ [$-0.000$, $+0.004$] \\
SD-Turbo, $t$ index $15$ & ukiyo\_e & $+0.125$ [$+0.103$, $+0.145$] & $+0.007$ [$+0.003$, $+0.011$] & $+0.016$ [$+0.011$, $+0.022$] \\
 & van\_gogh & $+0.132$ [$+0.092$, $+0.175$] & $+0.002$ [$-0.003$, $+0.007$] & $-0.004$ [$-0.009$, $-0.000$] \\
SD-Turbo, $t$ index $35$ & ukiyo\_e & $+0.015$ [$+0.010$, $+0.020$] & $-0.002$ [$-0.003$, $+0.000$] & $-0.000$ [$-0.002$, $+0.002$] \\
 & van\_gogh & $+0.017$ [$+0.011$, $+0.024$] & $-0.003$ [$-0.005$, $-0.002$] & $+0.000$ [$-0.002$, $+0.003$] \\
Aria Everyday Activities & ukiyo\_e & $+0.063$ [$+0.048$, $+0.081$] & $-0.006$ [$-0.009$, $-0.004$] & $+0.005$ [$-0.000$, $+0.009$] \\
 & van\_gogh & $+0.083$ [$+0.070$, $+0.098$] & $-0.009$ [$-0.011$, $-0.006$] & $-0.003$ [$-0.010$, $+0.002$] \\
\bottomrule
\end{tabular}}
\caption{The high-band main effect is positive in all $14$
cells; the low-band effect is near zero throughout. Factorial main effects on raw CSD gain, mean [clip-cluster
bootstrap $95\%$ CI], $n = 10$ clips or recordings per cell,
$K{=}6$. The SD-Turbo $t$ index $35$ teacher sits close to the
source (its own gain is near zero), so its effects are small in
absolute terms; the $t$ index $15$ teacher preserves little
content (teacher DINO $0.08$).}
\label{tab:supp-bandfactorial}
\end{table*}

The ordering is the same in every setting. The high-band main
effect is positive in all $14$ cells, with lower interval bound
at least $+0.010$; the low-band main effect is near zero
throughout (its interval excludes zero in four cells, each with
$|\text{effect}| \le 0.009$, two of them negative on the Aria
recordings), and interactions are small. On the Aria recordings
the retention levels (ratio of means) are $0.403$ (ukiyo\_e) and
$0.470$ (van\_gogh) for anchor sampling against $0.085$ and
$0.052$ for the chained low band.

\section{Causal streaming latency}
\label{sec:supp-latency}

The main paper reports amortized per-frame cost. This section
measures the same production pipeline as a causal stream: frames
are fed one at a time, and each frame's latency runs from the
moment the frame is pulled to the moment its stylized output is
emitted, with a CUDA synchronisation on both sides of every
frame. The timed function is the production streaming loop
itself (teacher run at keyframes, anchor propagation otherwise),
run on the confirmatory set ($10$ clips, both styles, $60$
frames, $K{=}6$, $3$ repeats, batch $1$). The headline population
is frames $6$--$59$, where the keyframe share is exactly $1/6$.

\begin{table*}[t]
\centering
\small
\begin{tabular}{lccccc}
\toprule
Host & All frames & Median & p95 & Keyframe & Non-keyframe \\
\midrule
RTX 5080 Laptop & $52.56 \pm 0.25$ & $19.69$ & $215.4$ & $212.94 \pm 1.03$ & $20.49 \pm 0.13$ \\
RTX 4090 & $45.24 \pm 0.09$ & $22.37$ & $159.2$ & $159.81 \pm 0.67$ & $22.33 \pm 0.20$ \\
\bottomrule
\end{tabular}
\caption{Causal latency is bimodal: about $20$ ms on skipped
frames, about $213$ ms at keyframes. Causal per-frame latency in ms (mean $\pm$ SD over $3$
repeats; $3{,}240$ frames per host). Sustained causal throughput
is $19.12 \pm 0.07$ frames/s on the RTX 5080 Laptop and
$22.32 \pm 0.01$ frames/s on the RTX 4090 (a second 4090 session
gave $45.14 \pm 0.81$ ms and $22.40$ frames/s). Frame $0$, which
needs no flow, takes $193.45 \pm 1.42$ ms on the RTX 5080 Laptop.
$16.9\%$ of frames exceed $33.3$ ms on the RTX 5080 Laptop: every
keyframe and $0.3\%$ of non-keyframes.}
\label{tab:supp-latency}
\end{table*}

The latency is bimodal by construction: a non-keyframe costs
about $20$ ms, a keyframe about $213$ ms on the timing host of
the main paper. The main paper's $51.12$ ms/frame came from this
same causal loop; its frame mask held a $15.9\%$ keyframe share
instead of $1/6$, and reweighting its own keyframe and
non-keyframe means ($214.91$ / $20.09$ ms) to a $1/6$ share gives
$52.56$ ms, so the two numbers differ by keyframe counting, not
by measurement kind. The per-frame synchronisation itself adds
$0.08$ ms/frame. Under paced arrival at $15$ frames/s on the
RTX 5080 Laptop the stream kept up in two of three repeats; the
third showed an unexplained stretch of slow keyframes and a
queue built up (mean latency $125.6$ ms over the three repeats).

Paced arrival was measured at three rates in a follow-up
(R12-A protocol, same harness). At $20$ frames/s the stream
falls behind in every run: closed-loop throughput is $19.2$
frames/s, so the backlog grows by about $170$ ms over each
$60$-frame stream. At $10$ and $15$ frames/s the outcome depends
on a slow state of the host that we could not identify: once a
process enters it, keyframes take about $380$--$395$ ms and
non-keyframes about $48$--$50$ ms (mean service $104$ ms/frame),
and it stays in that state until the process ends. The state
appeared in all four processes run at $10$ frames/s and in two
of three at $15$ frames/s, and in no closed-loop or $20$
frames/s process. In the fast state the stream keeps up at both
rates (mean latency about $79$ ms at $10$ frames/s and $103$ ms
at $15$ frames/s, final backlog about $0$). GPU utilisation is
low while the slow state lasts, and power and temperature
overlap between the states; a laptop power-management effect
is one candidate explanation and was not tested. The practical
reading is that the pipeline sustains $15$ frames/s arrival on
this laptop when the host stays in its fast state, and does not
sustain $20$ frames/s.

Feathering adds a fixed delay on top of this: the last three
frames before a keyframe are finalised only when that keyframe
has been computed, a wait of $3$ input frames, i.e.\ $157.7$ ms
at the anchor-sampling cadence ($162.7$ ms measured with
feathering on). The feather blend itself costs $8.72 \pm 0.20$
ms per keyframe ($2.89$ ms per finalised frame); AS + feather
averages $54.24 \pm 0.16$ ms/frame ($18.58$ frames/s) on the
RTX 5080 Laptop.

Two deployment notes follow from the bimodality. A player that
buffers $K$ frames (one keyframe interval, $315$ ms at the
measured cadence) absorbs the keyframe spike entirely, at the
cost of that much added display latency; feathering's
$3$-frame buffer is half of that. Alternatively the teacher can
run asynchronously: while a keyframe is being computed, the
stream keeps emitting frames warped from the previous anchor,
and the new anchor takes over when ready. This keeps the output
cadence flat at the non-keyframe cost and adds up to one
interval of anchor staleness instead of display delay. The
timed loop above is the synchronous pipeline; the asynchronous
policy's quality cost is measured in
\cref{sec:supp-ksweep}.

\section{Refresh interval, and the asynchronous policy}
\label{sec:supp-ksweep}

The main paper prices anchor sampling at $K{=}6$ and $K{=}9$.
\Cref{tab:supp-ksweep} maps the trade-off on the confirmatory
set for $K \in \{2, 3, 4, 6, 9, 12\}$: raw CSD gain falls as
the interval grows, the anchor's advantage over the chained
low-band baseline is positive at every interval (smallest at
$K{=}2$, where the chain is refreshed almost as often as the
anchor), and the boundary jump grows only mildly. The cost
column combines the directly measured pipeline means at
$K{=}3, 6, 9$ with the component model of
\cref{sec:supp-bandrefresh} at the other intervals (teacher
$222.11$ ms/frame, propagation $19.83$ ms/frame); the model
sits $1$--$2$ ms/frame above the measured means where both
exist.

\begin{table*}[t]
\centering
\small
\begin{tabular}{lccccccc}
\toprule
 & \multicolumn{2}{c}{AS raw gain} & \multicolumn{2}{c}{AS $-$ chained low-band} & \multicolumn{2}{c}{AS boundary T\_LPIPS} & ms/ \\
$K$ & ukiyo\_e & van\_gogh & ukiyo\_e & van\_gogh & ukiyo\_e & van\_gogh & frame \\
\midrule
$2$ & $0.209$ & $0.177$ & $+0.029$ [$+0.024$, $+0.037$] & $+0.045$ [$+0.037$, $+0.053$] & $0.204$ & $0.210$ & $121.0^{*}$ \\
$3$ & $0.183$ & $0.170$ & $+0.041$ [$+0.031$, $+0.053$] & $+0.065$ [$+0.054$, $+0.077$] & $0.214$ & $0.218$ & $87.27$ \\
$4$ & $0.171$ & $0.168$ & $+0.048$ [$+0.039$, $+0.060$] & $+0.073$ [$+0.059$, $+0.089$] & $0.220$ & $0.224$ & $70.4^{*}$ \\
$6$ & $0.140$ & $0.156$ & $+0.049$ [$+0.038$, $+0.061$] & $+0.080$ [$+0.063$, $+0.098$] & $0.226$ & $0.229$ & $51.12$ \\
$9$ & $0.121$ & $0.147$ & $+0.043$ [$+0.035$, $+0.055$] & $+0.079$ [$+0.058$, $+0.103$] & $0.230$ & $0.233$ & $40.83$ \\
$12$ & $0.100$ & $0.131$ & $+0.043$ [$+0.033$, $+0.055$] & $+0.078$ [$+0.056$, $+0.102$] & $0.230$ & $0.233$ & $36.7^{*}$ \\
\bottomrule
\end{tabular}
\caption{The compute--style curve in the refresh interval
(confirmatory set, $n = 10$ clips; differences are paired by
clip, $95\%$ clip-bootstrap intervals). Anchor sampling keeps
more of the teacher's style than the chained baseline at every
interval, at a boundary jump that grows from $0.20$ to $0.23$.
$^{*}$Component-model estimate; the other costs are measured
pipeline means.}
\label{tab:supp-ksweep}
\end{table*}

\paragraph{The asynchronous policy, measured.}
We also ran the asynchronous variant of
\cref{sec:supp-latency}: the teacher for keyframe $j$ is
computed during the previous interval, so each interval is
displayed from the previous keyframe's anchor and anchor ages
run $K{+}1$ to $2K$ instead of $1$ to $K$ ($K{=}6$,
confirmatory set). Taking the teacher off the critical path is
not free. Raw gain falls by $0.067$ [$0.054$, $0.081$] on
ukiyo\_e and $0.043$ [$0.017$, $0.067$] on van\_gogh against
the synchronous pipeline (retention $0.266$ / $0.451$ against
$0.506$ / $0.444$; the van\_gogh retention difference is inside
the per-frame estimator's noise, the raw-gain difference is
not). Two things improve: content similarity rises (DINO
$0.901$ / $0.890$ against $0.837$ / $0.822$), because a staler
anchor drifts less from the source, and total T\_LPIPS falls
($0.111$ / $0.112$ against $0.142$ / $0.141$). The
anchor-over-chain ordering survives the policy change:
asynchronous AS exceeds asynchronous chained low-band by
$+0.024$ [$+0.017$, $+0.032$] (ukiyo\_e) and $+0.055$
[$+0.035$, $+0.078$] (van\_gogh) in raw gain. In exchange, the
per-frame service time on the display path is the non-keyframe
cost alone ($20.49 \pm 0.13$ ms on the RTX 5080 Laptop,
\cref{sec:supp-latency}): no frame waits for the teacher, so
the $213$ ms keyframe stall of the synchronous loop disappears
from the output cadence. One qualification: the quality numbers
above measure the stale-anchor policy's output, and the service
figure is the measured non-keyframe cost; running the teacher
concurrently with propagation on one GPU was not benchmarked,
so contention could in principle eat into that margin.

\section{A second flow estimator}
\label{sec:supp-raftlarge}

Every result so far uses RAFT-small for flow and confidence.
This section replaces it with RAFT-large (the torchvision
$C\_T\_SKHT\_V2$ weights, run at $448 \times 768$ with $12$
updates, against RAFT-small's $224 \times 384$ with $4$) and
repeats the confirmatory comparison on the same stored teacher
frames, so only the adjacent flow and its confidence mask
change. The confidence definition is unchanged: forward--backward
error on the frame grid, with the same $3$ px cutoff, $1.5$ px
scale and $0.5$ visibility threshold in frame pixels. The two
estimators' backward flows differ by $1.24$ px mean endpoint
distance ($0.67$ px where both mark the pixel visible;
visibility agreement $0.90$). The evaluator first reproduced
the stored RAFT-small numbers exactly (retention $0.506$ /
$0.444$, T\_LPIPS $0.1415$, boundary $0.2275$, DINO $0.829$).

\begin{table*}[t]
\centering
\small
\begin{tabular}{llcc}
\toprule
Style & Quantity (raw CSD gain) & RAFT-small & RAFT-large \\
\midrule
ukiyo\_e & AS $-$ chained low-band & $+0.049$ [$+0.039$, $+0.060$] & $+0.051$ [$+0.036$, $+0.066$] \\
 & High-band main effect & $+0.048$ [$+0.039$, $+0.059$] & $+0.053$ [$+0.039$, $+0.066$] \\
 & Low-band main effect & $+0.001$ [$-0.003$, $+0.005$] & $-0.002$ [$-0.007$, $+0.004$] \\
 & Interaction & $-0.003$ [$-0.007$, $+0.002$] & $-0.002$ [$-0.005$, $+0.001$] \\
van\_gogh & AS $-$ chained low-band & $+0.080$ [$+0.064$, $+0.097$] & $+0.070$ [$+0.054$, $+0.088$] \\
 & High-band main effect & $+0.084$ [$+0.069$, $+0.100$] & $+0.076$ [$+0.062$, $+0.091$] \\
 & Low-band main effect & $-0.004$ [$-0.008$, $+0.001$] & $-0.006$ [$-0.011$, $-0.000$] \\
 & Interaction & $-0.005$ [$-0.008$, $-0.003$] & $+0.003$ [$-0.001$, $+0.006$] \\
\bottomrule
\end{tabular}
\caption{The anchor-over-chain conclusion does not depend on
the flow estimator. Confirmatory comparison under two flow estimators
($10$ clips, $K{=}6$, clip-cluster bootstrap $95\%$ intervals).
AS $-$ chained low-band is positive in $10$ of $10$ clips in
both styles under RAFT-large.}
\label{tab:supp-raftlarge}
\end{table*}

The conclusion is unchanged: AS $-$ chained low-band stays
positive with intervals excluding zero in both styles, and the
paired change against RAFT-small includes zero in both styles
($+0.002$ [$-0.009$, $+0.014$] ukiyo\_e; $-0.010$ [$-0.024$,
$+0.002$] van\_gogh). The levels move: the fallback fraction
falls from $0.268$ to $0.160$, AS retention rises to $0.661$ /
$0.575$, AS T\_LPIPS (stored-estimator ruler) falls to $0.130$
total and $0.219$ at boundaries, and DINO to source falls for
both methods (AS $0.829 \to 0.780$). The swap changes the
network, its training data, the input resolution and the
iteration count together, so these level changes are not
attributed to any one of them; a RAFT-large run at
$224 \times 384$ was not done.

The quality result above is not free. Flow and confidence with
RAFT-large take $118.91 \pm 0.13$ ms per frame against
$16.65 \pm 0.62$ ms for RAFT-small, and the full causal
pipeline with RAFT-large averages $156.54 \pm 0.17$ ms/frame
($6.50$ frames/s) on the RTX 5080 Laptop, against $52.26 \pm
0.12$ ms in the same session with RAFT-small. At that cost the
pipeline is slower than even a $10$ frames/s arrival rate, so
the RAFT-large numbers answer the estimator question, not a
deployment configuration.

\section{A streaming diffusion system}
\label{sec:supp-streamdiffusion}

StreamDiffusion (official implementation, commit b623251;
SD-Turbo with TAESD, fp16, no TensorRT) is a streaming system
with a different mechanism: a text prompt restyles every frame,
with no teacher keyframes, no residual reuse and no optical
flow. The comparison is therefore at system level and unmatched.
Configuration was chosen on development clips only, under a
rule fixed in advance: among prompt--strength pairs whose
DINO to source is at least $0.60$, take the highest CSD gain.
Only the as-shipped strength, $t$ index list $[35, 45]$, met
the content floor on development clips (style-mean gain
$+0.006$ at DINO $0.721$, against anchor sampling's $+0.149$ at
$0.836$ on the same frames); it is the primary setting (SD35).
A stronger setting, $[25, 40]$ (SD25), whose development gain
was closest to anchor sampling's, was added as an exploratory
second point before any test frame was run. Both use one fixed
prompt per style. All $2{,}400$ test frames ($10$ clips, both
styles, $60$ frames at $768 \times 448$) were generated and
scored with the paper's evaluator; the stored AS scores were
reproduced exactly in the same run.

\begin{table*}[t]
\centering
\small
\begin{tabular}{lcccccc}
\toprule
 & \multicolumn{2}{c}{CSD gain} & \multicolumn{2}{c}{DINO} & \multicolumn{2}{c}{T\_LPIPS total} \\
Method & ukiyo\_e & van\_gogh & ukiyo\_e & van\_gogh & ukiyo\_e & van\_gogh \\
\midrule
StreamDiffusion SD35 & $+0.064$ & $-0.015$ & $0.645$ & $0.665$ & $0.195$ & $0.193$ \\
StreamDiffusion SD25 & $+0.257$ & $+0.130$ & $0.269$ & $0.307$ & $0.260$ & $0.234$ \\
Anchor sampling & $+0.141$ & $+0.156$ & $0.837$ & $0.822$ & $0.142$ & $0.141$ \\
Teacher & $+0.277$ & $+0.225$ & $0.640$ & $0.637$ & $0.251$ & $0.256$ \\
\bottomrule
\end{tabular}
\caption{A prompt-driven streaming system does not reach
anchor sampling's style and content scores together at either
strength tried. StreamDiffusion on the confirmatory set (clip means).
The boundary column is not shown: at the frames that are
keyframes for anchor sampling, SD35's T\_LPIPS is $0.197$ /
$0.193$, which has no boundary meaning for a system without
keyframes.}
\label{tab:supp-streamdiffusion}
\end{table*}

At the primary setting, StreamDiffusion sits below anchor
sampling on style and content together (paired differences,
style-averaged per clip: CSD gain $-0.123$ [$-0.153$, $-0.094$],
DINO $-0.175$ [$-0.290$, $-0.068$], T\_LPIPS $+0.052$ [$+0.027$,
$+0.074$]); its gain is $0.23$ of the teacher's on ukiyo\_e and
$-0.07$ on van\_gogh. The stronger SD25 reaches anchor
sampling's gain on ukiyo\_e only, at DINO $0.269$ / $0.307$,
where the scene content is largely replaced. On latency the
system is flat rather than bimodal: SD35 takes $41.2$ ms/frame
on the RTX 4090 (p95 $45.8$ ms; $24.3$ frames/s), against
$45.1$--$45.2$ ms for anchor sampling on the same host, plus a
one-frame stream-batch delay. On the RTX 5080 Laptop it takes
$106.3$ ms/frame ($9.4$ frames/s) against anchor sampling's
$52.6$ ms; that host ran with a fallback attention kernel (the
xformers kernel does not support the RTX 5080 in this
environment), so the cross-host gap is not attributed to the
model. Two scope notes: the CSD reference is a single style
image that the teacher is conditioned on and StreamDiffusion
never sees, which favours image-conditioned systems; and
TensorRT, which the project's own benchmarks use, was not run.

\section{Chained baselines at full strength, and oracle controls}
\label{sec:supp-chains}

The chained baseline of the main paper carries only the low
band into its keyframes. This section runs the chain family at
full strength on the confirmatory set ($K{=}6$, $n = 10$ clips):
the full-band chained residual (D0: at a keyframe the residual
is the teacher's, elsewhere $r = c \cdot \mathrm{warp}(r)$), the
same chain with a Gaussian fill on low-confidence pixels
(LOW-b), both chains and the paper's baseline with bicubic
instead of bilinear warping, a sharpness-matched version of
anchor sampling itself, and an oracle control that hands the
chain anchor sampling's own radial spectrum. All differences
are paired, with clip-cluster bootstrap intervals (2,000 draws).

\begin{table*}[t]
\centering
\small
\begin{tabular}{lcccccc}
\toprule
& \multicolumn{3}{c}{ukiyo\_e} & \multicolumn{3}{c}{van\_gogh} \\
Method & raw gain & retention & boundary & raw gain & retention & boundary \\
\midrule
Anchor sampling & $0.140$ & $0.506$ & $0.226$ & $0.156$ & $0.444$ & $0.229$ \\
Chained low-band (paper) & $0.092$ & $0.332$ & $0.179$ & $0.075$ & $0.175$ & $0.181$ \\
Full-band chain (D0) & $0.106$ & $0.377$ & $0.340$ & $0.103$ & $0.219$ & $0.332$ \\
Chain with Gaussian fill (LOW-b) & $0.130$ & $0.469$ & $0.338$ & $0.126$ & $0.373$ & $0.328$ \\
D0, bicubic warp & $0.123$ & $0.441$ & $0.346$ & $0.128$ & $0.318$ & $0.339$ \\
LOW-b, bicubic warp & $0.147$ & $0.530$ & $0.344$ & $0.153$ & $0.455$ & $0.336$ \\
Chained low-band, bicubic & $0.107$ & $0.386$ & $0.179$ & $0.103$ & $0.263$ & $0.180$ \\
Spectrum-matched chain (oracle) & $0.131$ & $0.473$ & $0.220$ & $0.161$ & $0.476$ & $0.222$ \\
\bottomrule
\end{tabular}
\caption{The chain family on the confirmatory set. Raw gain is
the raw CSD gain, retention the per-frame CSD retention, and
boundary the boundary T\_LPIPS (lower is better). The full
metric set, including DINO, total and internal T\_LPIPS and
LPIPS to the teacher, is in the experiment record.}
\label{tab:supp-chains}
\end{table*}

Three findings. First, anchor sampling exceeds the full-band
chain in raw gain by $+0.035$ [$+0.023$, $+0.046$] on ukiyo\_e
and $+0.053$ [$+0.039$, $+0.068$] on van\_gogh, and the chain
with a Gaussian fill by $+0.010$ [$+0.005$, $+0.015$] and
$+0.030$ [$+0.024$, $+0.036$]; carrying the whole residual
instead of only the low band does not close the gap. The
full-band chains pay for their gain at the keyframes: their
boundary T\_LPIPS is $0.33$--$0.35$ against anchor sampling's
$0.23$, because the chained residual has drifted by the time
the keyframe resets it.

Second, the resampler matters and we report it rather than
choose the weaker one: bicubic warping raises every chain's
raw gain by $+0.015$ to $+0.028$. Anchor sampling still
exceeds the paper's baseline in bicubic form by $+0.034$ /
$+0.052$ and the full-band bicubic chain by $+0.017$ /
$+0.028$. The one exception is the bicubic chain with a
Gaussian fill and no keyframe blend: it scores above anchor
sampling on ukiyo\_e by $0.007$ [$0.002$, $0.013$] and is not
separated from it on van\_gogh ($+0.002$ [$-0.002$, $+0.007$]
for anchor sampling). Two qualifications apply. That chain
resets to the exact teacher frame at every keyframe, and one
of the eleven scored frames (frame $30$) is such a keyframe;
leaving that frame out, in an exploratory split, gives
$+0.003$ / $+0.013$ for anchor sampling. And its keyframe
jump remains half again as large as anchor sampling's
(boundary T\_LPIPS $0.344$ / $0.336$ against $0.226$ /
$0.229$), with content similarity equal (DINO differences
$+0.002$ / $+0.001$). The bicubic chains are also sharper
than the teacher itself ($1.37\times$), so part of their
gain may be sharpening; this was not separated.

The same comparison also isolates the keyframe construction.
The exact-reset chain's keyframes are exact teacher frames,
where anchor sampling displays a confidence-weighted blend.
Changing only that --- giving the bicubic Gaussian-filled
chain anchor sampling's blended keyframe construction, with
the same flows, fill and resampling --- produces frames
pixel-identical to the bicubic chained low-band baseline of
\cref{tab:supp-chains} on all $1{,}200$ frames: under a matched
keyframe construction, the two are the same method. The
keyframe construction alone therefore accounts for the
exact-reset chain's style parity (replacing the reset with
the blend lowers its raw gain by $0.041$ [$0.031$, $0.053$] on
ukiyo\_e and $0.050$ [$0.039$, $0.062$] on van\_gogh, paired)
and for its boundary penalty (boundary T\_LPIPS lower by
$0.165$ [$0.142$, $0.194$] and $0.156$ [$0.143$, $0.169$]).
Under the matched construction the comparison is the
tabulated one: anchor sampling leads the chain by $0.034$ and
$0.052$ of raw gain, while the chain leads on the boundary
jump ($0.179$ and $0.180$ against $0.226$ and $0.229$) and on
DINO.

Third, two controls locate the mechanism. Blurring anchor
sampling to the teacher's sharpness leaves its advantage
almost untouched: the blurred version exceeds the chained
low-band baseline by $+0.048$ / $+0.079$ in raw gain, against
$+0.049$ / $+0.080$ unblurred, so the advantage is not a
sharpness artefact. Matching the chain's high-band radial
spectrum to anchor sampling's own spectrum, frame by frame,
closes $79.9\%$ [$72.2$, $88.7$] of the raw-gain gap on
ukiyo\_e and $106.7\%$ [$95.6$, $120.2$] on van\_gogh, where
amplitude matching alone closes $45.6\%$ / $82.7\%$. A small
remainder survives on ukiyo\_e (anchor sampling minus the
spectrum-matched chain $+0.010$ [$+0.005$, $+0.016$]); on
van\_gogh the difference is not separated from zero
($-0.005$ [$-0.015$, $+0.004$]). The spectrum-matched chain
is an oracle: it reads the spectrum from anchor sampling's
own output, so it is not a runnable method. What it shows is
that the anchor's advantage over the chain is, in the main,
the high-band spectrum that iterated warping loses ---
which is the quantity the anchor exists to preserve.

\section{Master table for the confirmatory set}
\label{sec:supp-master}

\begin{figure}[t]
\centering
\includegraphics[width=\columnwidth]{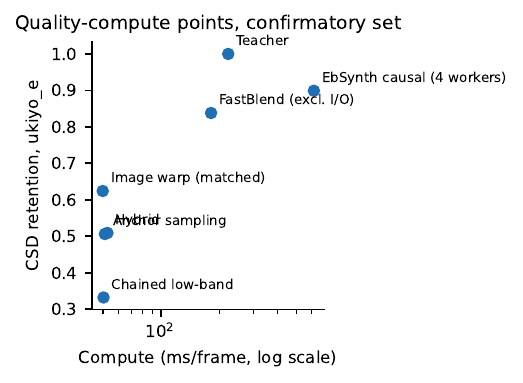}
\caption{Quality--compute points from \cref{tab:supp-master}
(ukiyo\_e retention; EbSynth at its 4-worker throughput).
No method dominates: the offline and CPU points sit higher and
far to the right; the causal GPU points cluster near
$50$ ms/frame.}
\label{fig:supp-pareto}
\end{figure}

\Cref{tab:supp-master} gathers the confirmatory-set numbers
that the main paper spreads over several tables, and
\cref{fig:supp-pareto} plots the quality--compute points. Costs
are amortized ms/frame on the RTX 5080 Laptop unless a host is
named; the buffer column is the number of future frames a method
must see before a frame can be emitted.

\begin{table*}[t]
\centering
\small
\resizebox{\textwidth}{!}{%
\begin{tabular}{lcccccccccc}
\toprule
Method & ms/frame & Buffer & \multicolumn{2}{c}{Retention} & \multicolumn{2}{c}{Raw gain} & DINO & T\_LPIPS & \multicolumn{2}{c}{LPIPS to teacher} \\
 & & (frames) & ukiyo\_e & van\_gogh & ukiyo\_e & van\_gogh & & total / boundary & whole & flow-valid \\
\midrule
Teacher & $222.11$ & $0$ & $1.000$ & $1.000$ & $+0.277$ & $+0.225$ & $0.638$ & $0.254$ / $0.254$ & $0.000$ & $0.000$ \\
Anchor sampling & $51.12$ & $0$ & $0.506$ & $0.444$ & $+0.141$ & $+0.156$ & $0.829$ & $0.142$ / $0.228$ & $0.267$ & $0.300$ \\
Hybrid & $52.66$ & $0$ & $0.509$ & $0.472$ & $+0.141$ & $+0.162$ & --- & --- & --- & --- \\
Chained low-band & $50.29$ & $0$ & $0.332$ & $0.174$ & $+0.092$ & $+0.075$ & $0.913$ & $0.112$ / $0.180$ & $0.264$ & $0.288$ \\
Image warp (matched) & $\approx 49.8$ & $0$ & $0.624$ & $0.476$ & --- & --- & $0.760$ & $0.110$ / $0.253$ & $0.276$ & --- \\
FastBlend & $181$ excl.\ I/O & $6$ & $0.838$ & $0.719$ & --- & --- & $0.690$ & $0.093$ / $0.119$ & $0.206$ & --- \\
EbSynth (causal, CPU) & $616$ at $4$ workers & $0$ & $0.899$ & $0.842$ & --- & --- & $0.628$ & $0.127$ / $0.316$ & $0.218$ & --- \\
StreamDiffusion SD35 & $41.2$ (RTX 4090) & $1$ & --- & --- & $+0.064$ & $-0.015$ & $0.655$ & $0.194$ / --- & --- & --- \\
\bottomrule
\end{tabular}}
\caption{Confirmatory set, $K{=}6$. LPIPS to teacher is split
by the region it is computed on: whole-frame values and
flow-valid-region values sit in separate columns, because the
two are not interchangeable (anchor sampling: $0.267$ whole
frame against $0.300$ flow-valid). A dash means that region was
not recorded for that method. StreamDiffusion's raw gain is not
a retention: its conditioning (a text prompt) differs from the
teacher's (a style image), see \cref{sec:supp-streamdiffusion}.
FastBlend's end-to-end cost including file I/O is
$\approx 243$ ms/frame (\cref{sec:supp-ebsynth}).}
\label{tab:supp-master}
\end{table*}

\begin{figure*}[t]
\centering
\includegraphics[width=0.62\textwidth]{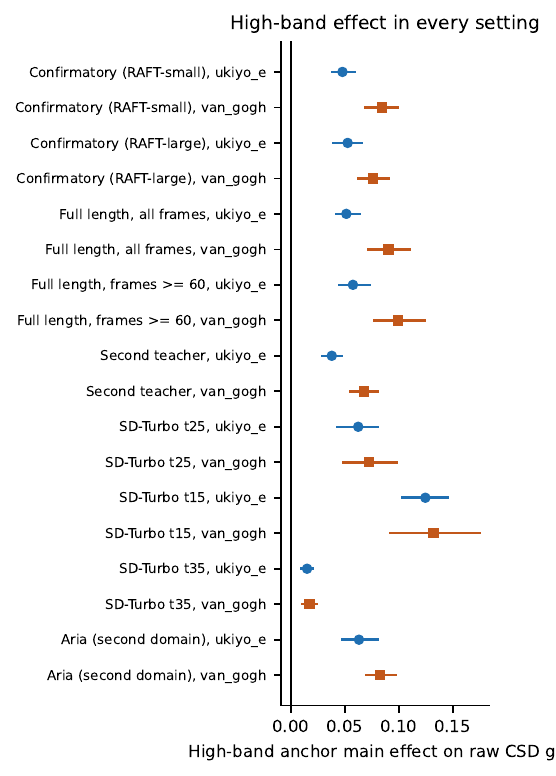}
\caption{The high-band anchor main effect on raw CSD gain is
positive in every setting in which the factorial was run:
two flow estimators, full-length sequences, three teacher
configurations, and a second, egocentric domain. The low-band
effect is near zero throughout (\cref{sec:supp-bandfactorial}).
Circles are ukiyo\_e, squares van\_gogh; bars are clip-cluster
bootstrap $95\%$ intervals.}
\label{fig:supp-forest}
\end{figure*}

\section{Position among resampling and keyframe propagation}
\label{sec:supp-position}

Three things were already established before this study.
Repeated bilinear resampling attenuates high frequencies; this
is classical sampling theory, and it is why chained warping
loses detail. Keyframe propagation itself is an old idea in
video stylization (EbSynth and its predecessors) and in
efficient video recognition (keyframe feature propagation).
And any keyframe scheme concentrates its error at refresh
boundaries. What this study adds is measurement, on a modern
diffusion teacher: how much of the style a sparse-keyframe
system loses is attributable to the transport rule rather than
to the teacher (the factorial and the amplitude- and
spectrum-matched controls), how that attribution survives
changes of teacher, flow estimator, sequence length and domain
(\cref{fig:supp-forest}), where the temporal error sits
(boundary concentration, anchor-age decay), and what the usual
remedy for the boundary error costs in content and in human
preference (\cref{sec:supp-vehicle,sec:supp-human}). The
contribution is the quantification and its replication, not the
observation that resampling blurs.

\section{Reading the style metrics}
\label{sec:supp-metricreading}

Retention normalises each frame's CSD gain by the teacher's
gain on the same frame. Two consequences deserve emphasis.
First, retention is a ratio of CSD scores, not a perceptual
quantity: $0.5$ retention does not mean a viewer perceives half
the style. Second, the per-frame ratio is unstable when the
teacher's gain on a frame is near zero, which is why the paper
also reports raw gain (an unnormalised difference) and, in
\cref{sec:supp-r5}, a ratio-of-means estimator. On the
confirmatory set the two estimators differ mainly on van\_gogh
(AS $0.444$ per-frame against $0.691$ ratio-of-means); every
ordering the paper relies on (anchor over chain, the factorial
effects, the replication across settings) is stated on raw gain
or holds under both estimators, while exact retention levels
should be read as estimator-dependent. The two primary styles
were chosen because CSD separates the teacher from the source
on them; on watercolor and ink sketch the teacher's CSD gain is
small and retention is not informative there, so those styles
are reported in raw terms in \cref{sec:supp-styles}. CSD is
computed on the whole frame; the flow-valid-region analyses in
\cref{sec:supp-csd-abs} use the same frames with the fallback
region filled by each method's own output, so region gains need
not average to the whole-frame gain.

\section{Human preferences and the automatic metrics}
\label{sec:supp-human}

\begin{figure*}[t]
\centering
\includegraphics[width=0.95\textwidth]{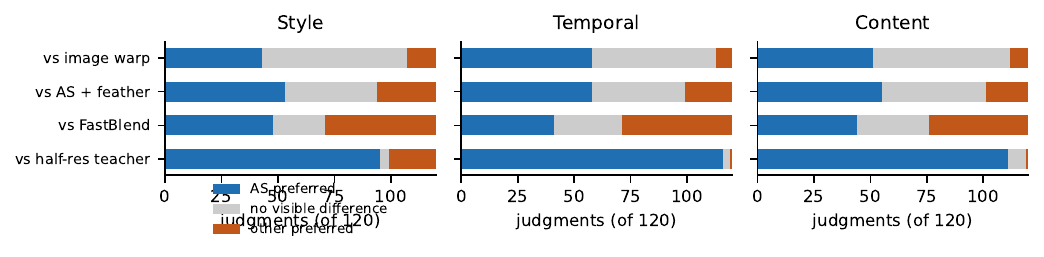}
\caption{Human preferences (all $15$ raters, wins--ties--losses
from anchor sampling's perspective). Anchor sampling is
preferred over image warp and the half-resolution teacher,
roughly tied with FastBlend, and preferred over its own
feathered variant despite feathering's better boundary metric.}
\label{fig:supp-human}
\end{figure*}

Protocol. Raters compared pairs of six-second clips (the same
$60$ frames the automatic metrics score) side by side and, for
each pair, chose which clip was better on three axes (style,
temporal smoothness, content preservation) or reported no
visible difference. The left/right assignment was randomised
per pair; playback had to finish before the choices unlocked.
Each rater judged $32$ pairs drawn from the four comparisons
below; the operating points shown were the standard ones
($\alpha = 1$), not the content-matched variants of
\cref{sec:supp-matched}. Fifteen raters completed the study;
\cref{fig:supp-human} shows all $120$ judgments per axis.

Where the metrics and the raters disagree. On the two
comparisons where CSD and CLIP rank a method above anchor
sampling, the raters rank anchor sampling first: image warp has
the higher CSD retention ($0.66$ / $0.62$ against $0.51$ /
$0.44$) and the higher CLIP score, yet raters prefer anchor
sampling on style $77\%$ of non-tied judgments. Feathering
raises retention ($0.62$ / $0.60$) and halves the boundary
T\_LPIPS, yet raters prefer plain anchor sampling on style
($67\%$) and on temporal smoothness ($73\%$). The likely reading
is that the automatic style scores reward the amount of stylized
texture, while raters also penalise the artifacts each method
adds: image warp's stretching, and feathering's early import of
the next keyframe's content (the failure case in
\cref{sec:supp-feather-fail}), which a boundary-aligned flicker
metric does not see. We therefore treat CSD retention as a
measure of style transfer relative to the teacher, not as a
proxy for preference, and we report the human results wherever
a claim is perceptual.

\section{Baseline comparison at a glance}
\label{sec:supp-basematrix}

\begin{table*}[t]
\centering
\small
\resizebox{\textwidth}{!}{%
\begin{tabular}{llllll}
\toprule
Method & Conditioning & Future frames & Hardware & Backend & Timing includes \\
\midrule
Anchor sampling & style image (via teacher) & $0$ & RTX 5080 Laptop / 4090 & PyTorch & compute only \\
FastBlend & style image (same keyframes) & up to $6$ & RTX 4090 & DiffSynth-Studio & $181$ ms excl.\ I/O; $\approx 243$ incl. \\
EbSynth & style image (same keyframes) & $0$ (causal form) & CPU (i9-13900KF) & official CPU build & end-to-end incl.\ I/O \\
StreamDiffusion & text prompt & $1$ (stream batch) & RTX 4090 / 5080 Laptop & PyTorch, no TensorRT & compute only \\
Teacher & style image & $0$ & RTX 5080 Laptop & PyTorch & compute only \\
\bottomrule
\end{tabular}}
\caption{What differs between the systems compared in
\cref{tab:supp-master}. No two rows share every column, which is
why the paper's claims are stated per axis (causality, compute,
quality) rather than as a single ranking. EbSynth's official
repository also ships a CUDA backend; the controlled timing in
\cref{sec:supp-ebsynth} used the CPU build, and its
$0.62$--$1.36$ s/frame is $14$--$30\times$ anchor sampling's
$45.1$ ms/frame on the RTX 4090. The CUDA backend has since
been timed at $0.55$ s/frame.}
\label{tab:supp-basematrix}
\end{table*}

\section{Behaviour at scene cuts}
\label{sec:supp-scenecuts}

The implementation has no scene-cut detector. At a cut, the
composed flow between the last pre-cut frame and the first
post-cut frame is mostly invalid, the confidence falls below
threshold on most pixels, and the fallback (the chained
low-band estimate, itself carried across the cut) fills much of
the frame until the next keyframe, at most $K{-}1$ frames later;
the residual state is fully reset at that keyframe, because a
keyframe's residual is computed from the current frame alone.
A deployment can force a keyframe at a detected cut and remove
even that interval; we did not evaluate cut handling
quantitatively, and the test sets contain no cuts.

\section{Configuration details}
\label{sec:supp-config}

Settings not stated in the main paper. The teacher is SDXL-Turbo
with the small canny ControlNet variant and IP-Adapter (ViT-H);
it runs a $4$-step schedule at guidance scale $0$ with img2img
strength $0.65$ ($2$ denoising steps) and seed $20261104$. The
second teacher of \cref{sec:supp-teacher2} is SD-Turbo (SD~2.1).
RAFT-small runs at $224 \times 384$ input resolution with $4$
recurrent updates, its flows resized to $768 \times 448$. LPIPS
uses the AlexNet backbone; the T\_LPIPS evaluator averages the
spatial LPIPS map over pixels with confidence $\ge 0.5$ whose
sampled source position is valid. FastBlend is DiffSynth-Studio
v1.1.9 in interpolation mode with its default configuration.
EbSynth is the official source build (commit 2f5c97c) on CPU with
$8$ OpenMP threads. Stored quality frames were generated on an
RTX 4090 workstation; the main paper's timing protocol ran on an
RTX 5080 Laptop GPU, and the same-GPU comparison and EbSynth CPU
runs used an i9-13900KF ($32$ threads). Repeatability of the
timing protocol across saved repeats: at $K{=}6$, AS $51.12 \pm
0.23$, Hybrid $52.66 \pm 0.26$, chained low-band $50.29 \pm 0.31$
ms/frame (mean $\pm$ SD); at $K{=}9$, AS $40.83 \pm 0.15$
ms/frame; no repeat SD exceeds $0.31$ ms/frame.

\section{Stratification by motion and fallback fraction}
\label{sec:supp-strat}

\Cref{tab:supp-strat} gives the per-bin values behind the
stratification summary in the main paper ($100$ keyframe
intervals per style, terciles cut once over all intervals).

\begin{table*}[t]
\centering
\small
\begin{tabular}{llcccc}
\toprule
 & & \multicolumn{2}{c}{$\mathrm{AS} - \text{chained low-band}$ raw CSD gain} & \multicolumn{2}{c}{$(\text{AS + feather}) - \mathrm{AS}$ boundary T\_LPIPS} \\
\cmidrule(lr){3-4}\cmidrule(lr){5-6}
Covariate & Bin & ukiyo\_e & van\_gogh & ukiyo\_e & van\_gogh \\
\midrule
Motion & low  & $+0.055$ [$+0.041$, $+0.069$] & $+0.103$ [$+0.081$, $+0.129$] & $-0.097$ [$-0.125$, $-0.074$] & $-0.104$ [$-0.129$, $-0.081$] \\
 & mid  & $+0.057$ [$+0.041$, $+0.075$] & $+0.085$ [$+0.061$, $+0.112$] & $-0.112$ [$-0.133$, $-0.091$] & $-0.115$ [$-0.137$, $-0.094$] \\
 & high & $+0.034$ [$+0.020$, $+0.046$] & $+0.063$ [$+0.036$, $+0.090$] & $-0.125$ [$-0.165$, $-0.090$] & $-0.125$ [$-0.164$, $-0.091$] \\
\midrule
Fallback & low  & $+0.064$ [$+0.048$, $+0.080$] & $+0.118$ [$+0.097$, $+0.141$] & $-0.084$ [$-0.107$, $-0.068$] & $-0.094$ [$-0.120$, $-0.073$] \\
 & mid  & $+0.055$ [$+0.047$, $+0.065$] & $+0.092$ [$+0.073$, $+0.115$] & $-0.120$ [$-0.159$, $-0.085$] & $-0.122$ [$-0.157$, $-0.092$] \\
 & high & $+0.036$ [$+0.028$, $+0.045$] & $+0.054$ [$+0.038$, $+0.072$] & $-0.123$ [$-0.158$, $-0.093$] & $-0.122$ [$-0.156$, $-0.092$] \\
\bottomrule
\end{tabular}
\caption{Stratification on the confirmatory set ($K{=}6$).
Intervals are clip-cluster bootstrap intervals ($2{,}000$
resamples). AS boundary T\_LPIPS itself varies little across
bins ($0.203$--$0.233$, ukiyo\_e). The per-interval gain
correlates with fallback fraction at Spearman $-0.41$ [$-0.70$,
$+0.01$] (ukiyo\_e) and $-0.64$ [$-0.79$, $-0.35$] (van\_gogh).}
\label{tab:supp-strat}
\end{table*}

\section{Temporal diagnostics for all methods}
\label{sec:supp-temporal}

\Cref{tab:supp-temporal} completes the motion-normalized flicker
analysis of the main paper for all methods: unwarped LPIPS and
warp gain (unwarped minus warped) on the T\_LPIPS mask, with the
source sequence's own change ($0.0768$ over all transitions,
$0.0785$ at boundaries; $1{,}180$ transitions, $180$ boundaries)
as the normalization reference.

\begin{table*}[t]
\centering
\small
\begin{tabular}{lcccccc}
\toprule
 & \multicolumn{2}{c}{T\_LPIPS} & \multicolumn{2}{c}{$\div$ source change} & \multicolumn{2}{c}{Warp gain} \\
\cmidrule(lr){2-3}\cmidrule(lr){4-5}\cmidrule(lr){6-7}
Method & total & boundary & total & boundary & total & boundary \\
\midrule
Anchor sampling & $0.1415$ & $0.2275$ & $1.84$ & $2.90$ & $0.063$ & $0.058$ \\
Chained low-band & $0.1118$ & $0.1799$ & $1.46$ & $2.29$ & $0.067$ & $0.058$ \\
Image warp & $0.1260$ & $0.3290$ & $1.64$ & $4.19$ & $0.071$ & $0.007$ \\
Crossfade & $0.1298$ & $0.1294$ & $1.69$ & $1.65$ & $0.063$ & $0.051$ \\
AS + feather & $0.1244$ & $0.1101$ & $1.62$ & $1.40$ & $0.067$ & $0.060$ \\
FastBlend & $0.0930$ & $0.1191$ & $1.21$ & $1.52$ & $0.061$ & $0.045$ \\
\bottomrule
\end{tabular}
\caption{Temporal diagnostics on the confirmatory set ($K{=}6$,
both styles). Unwarped LPIPS (total / boundary): AS $0.205$ /
$0.286$, chained low-band $0.179$ / $0.238$, image warp $0.197$ /
$0.336$, crossfade $0.193$ / $0.180$, AS + feather $0.191$ /
$0.170$, FastBlend $0.154$ / $0.164$.}
\label{tab:supp-temporal}
\end{table*}

\section{Full-length sequences: remaining values}
\label{sec:supp-long}

Beyond the raw gains reported in the main paper, the full-length
extension ($62$--$91$ frames per clip; $176$ additional frames
per style) gives: Hybrid $-$ AS raw CSD gain beyond frame $60$ of
$-0.003$ [$-0.008$, $+0.002$] (ukiyo\_e) and $+0.005$ [$-0.000$,
$+0.010$] (van\_gogh); AS boundary T\_LPIPS of $0.227$ [$0.194$,
$0.260$], with boundaries carrying $1.54\times$ their uniform
share of T\_LPIPS mass; and an anchor-age curve of unchanged
shape (monotone decay for both methods in both windows, AS above
the chained estimate at every age, change in age-$5$ minus
age-$1$ drift at most $+0.007$ with intervals spanning zero).
\Cref{tab:supp-long} lists the retention levels. (Van\_gogh
retention beyond frame $60$ exceeds $0.9$ because the
teacher--source gap is small on several clips; the raw gain is
the robust quantity there.)

\begin{table*}[t]
\centering
\small
\begin{tabular}{lcc}
\toprule
Frames $\ge 60$ ($K{=}6$) & ukiyo\_e & van\_gogh \\
\midrule
$\mathrm{AS} - \text{chained low-band}$ retention & $+0.231$ [$+0.184$, $+0.282$] & $+0.577$ [$+0.364$, $+0.891$] \\
Retention: AS / Hybrid / chained low-band & $0.498$ / $0.491$ / $0.267$ & $0.964$ / $1.006$ / $0.387$ \\
AS T\_LPIPS total / boundary (both styles pooled) & \multicolumn{2}{c}{$0.155$ [$0.125$, $0.185$] / $0.227$ [$0.194$, $0.260$]} \\
\bottomrule
\end{tabular}
\caption{Full-length extension of the confirmatory set:
retention levels beyond frame $60$.}
\label{tab:supp-long}
\end{table*}

\section{Development clips and the sparse-regeneration baseline}
\label{sec:supp-devclips}

The development clips used for the design decisions reported
here (teacher calibration and the StreamDiffusion configuration
rule) are six DAVIS clips:
\emph{boat}, \emph{car-roundabout}, \emph{cows},
\emph{flamingo}, \emph{motocross-jump} and \emph{night-race}.
They are disjoint from the first test set, the confirmatory set
and the replication set. One baseline of the main paper's
Tab.~1 is described only there: sparse regeneration runs the
teacher again on the crops whose flow confidence is low and
composites them into the propagated frame; on the first test
set it costs $83.89$ ms/frame at $K{=}6$, between the one-step
teacher and anchor sampling, and it reaches higher retention
than anchor sampling there ($0.64$ against $0.59$ on ukiyo\_e
at $K{=}6$; $0.57$ against $0.51$ at $K{=}9$) at roughly $1.6$
times anchor sampling's cost. It is a costlier operating point
on the same trade-off, not a dominated one.

{
    \small
    \bibliographystyle{ieeenat_fullname}
    \bibliography{references}
}

\end{document}